\documentclass[10pt,twocolumn,letterpaper]{article}

\usepackage[pagenumbers]{wacv} 

\usepackage{times}
\usepackage{epsfig}
\usepackage{amssymb}
\usepackage{booktabs}
\usepackage{tabularx}
\usepackage{multirow}
\usepackage{subcaption}
\usepackage[export]{adjustbox}
\usepackage{float}
\usepackage{dashrule}
\usepackage{arydshln}
\usepackage[utf8x]{inputenc}
\usepackage[space]{grffile}
\usepackage{anyfontsize}
\usepackage{t1enc}
\usepackage{mathtools, nccmath}
\usepackage{graphicx}

\usepackage{amsmath,amsfonts}
\usepackage{algorithmic}
\usepackage{algorithm}
\usepackage{array}
\usepackage{textcomp}
\usepackage{stfloats}
\usepackage{url}
\usepackage{verbatim}
\usepackage{animate}
\usepackage{xcolor}
\usepackage{bbding}
\usepackage{tablefootnote}
\usepackage{xr-hyper}

\newlength \mywidth
\newlength \myw
\newlength \myww
\definecolor{myblue}{HTML}{006EAF}
\definecolor{myred}{HTML}{CC0000}

\def\ours{VLMDiff} 

\definecolor{wacvblue}{rgb}{0.21,0.49,0.74}
\usepackage[pagebackref,breaklinks,colorlinks,allcolors=wacvblue]{hyperref}

\def\wacvPaperID{2157} 
\def\confName{WACV}
\def\confYear{2026}

\title{VLMDiff: Leveraging Vision-Language Models for Multi-Class Anomaly Detection with Diffusion}

\author{Samet Hicsonmez, Abd El Rahman Shabayek, Djamila Aouada \\
University of Luxembourg\\
Luxembourg, Luxembourg\\
{\tt\small \{samet.hicsonmez,abdelrahman.shabayek,djamila.aouada\}@uni.lu}
}

\begin{document}
\maketitle
\begin{abstract}

Detecting visual anomalies in diverse, multi-class real-world images is a significant challenge. We introduce \ours, a novel unsupervised multi-class visual anomaly detection framework. It integrates a Latent Diffusion Model (LDM) with a Vision-Language Model (VLM) for enhanced anomaly localization and detection. Specifically, a pre-trained VLM with a simple prompt extracts detailed image descriptions, serving as additional conditioning for LDM training. Current diffusion-based methods rely on synthetic noise generation, limiting their generalization and requiring per-class model training, which hinders scalability. \ours, however, leverages VLMs to obtain normal captions without manual annotations or additional training. These descriptions condition the diffusion model, learning a robust normal image feature representation for multi-class anomaly detection. Our method achieves competitive performance, improving the pixel-level Per-Region-Overlap (PRO) metric by up to 25 points on the Real-IAD dataset and 8 points on the COCO-AD dataset, outperforming state-of-the-art diffusion-based approaches. Code is available at \url{https://github.com/giddyyupp/VLMDiff}.  
\end{abstract}    
\section{Introduction}
\label{sec:intro}

Visual Anomaly Detection (AD) involves identifying and localizing abnormal regions in images, with applications across various computer vision tasks. These include quality control in manufacturing~\cite{bergmann2018improving, visa, wang2024real} where detecting defects ensures product reliability, and autonomous driving~\cite{bogdoll2022anomaly} where identifying unseen obstacles is crucial for safety. The key challenge in visual AD lies in the scarcity of labeled abnormal samples—while normal cases are often well-documented, collecting and annotating anomalous instances is typically impractical due to their rarity and diversity~\cite{pang2021deep}. Consequently, unsupervised approaches that rely solely on normal examples are essential for real-world deployment, enabling robust AD without the need for exhaustive manual annotations~\cite{cui2023survey}.

\begin{figure}
\centering\resizebox{1.0\columnwidth}{!}{
\begin{tabular}{cc}
\includegraphics[height=0.5\columnwidth]{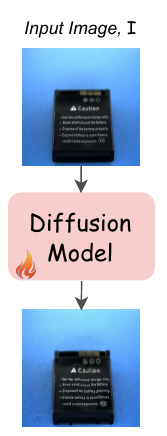} & 
\includegraphics[height=0.5\columnwidth]{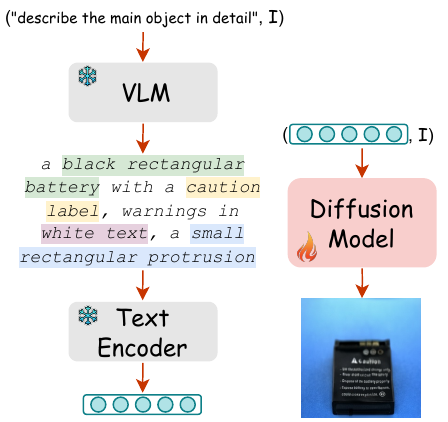} \\
(a)  & (b) \\
\end{tabular}}
\caption{Comparison of (a) current diffusion-based approaches and (b) our approach. Our method extracts textual descriptions from VLMs to guide the training of the underlying diffusion model. } 
\label{fig:teaser}
\vspace{-5mm}
\end{figure}

The primary approaches for visual AD can be categorized into augmentation-based~\cite{zavrtanik2021draem, liu2023simplenet,FAUAD}, embedding-based~\cite{lee2022cfa, gudovskiy2022cflow, lei2023pyramidflow}, reconstruction-based~\cite{tien2023revisiting, he2024mambaad, zhang2023exploring}, and hybrid approaches~\cite{uniad, zhang2023destseg} that combine multiple techniques. Among these, reconstruction-based methods emerge as a natural solution which train models to learn normal sample reconstructions.
During inference, when an anomalous sample is encountered, the model will attempt to reconstruct it as a normal sample, resulting in a higher reconstruction error compared to normal inputs. This reconstruction error is then used to identify anomalies.
Various architectures have been explored for this purpose, including autoencoders~\cite{akcay2019ganomaly}, masked vision transformers (ViTs)~\cite{zhang2023exploring}, and Mamba models~\cite{he2024mambaad}. 
Recently, diffusion-based~\cite{diff_first, ddpm, ldm} approaches have been proposed for AD~\cite{wyatt2022anoddpm, wolleb2022diffusion, zhang2023unsupervised, zhang2023diffusionad, he2024diffusion, livernochediffusion} as another line of reconstruction-based methods. 
However, first, their performance still lags behind the current state-of-the-art methods, especially on datasets with subtle defect types~\cite{wang2024real}, primarily due to weak guidance signals during training. Second, most of these models~\cite{zhang2023unsupervised, zhang2023diffusionad, Fucka_2024_ECCV, wolleb2022diffusion, wyatt2022anoddpm}, except for~\cite{he2024diffusion}, are trained on a per-class basis, making them impractical for large-scale datasets with multiple object categories. Last but not least, their pipeline relies on synthetic anomaly generation in the image space, which limits them to specific defect types and domains.

Another line of research~\cite{gu2024filo, zhouanomalyclip, zanella2024harnessing, cao2025adaclip} explores leveraging the strong zero-shot capabilities of vision-language models (VLMs) for visual AD. These methods rely on learning effective text prompts to generate accurate descriptions of both normal and anomalous cases. However, to learn prompts for anomalies, they typically require labeled auxiliary anomalous data from external datasets, which is often difficult to obtain in practice. Additionally, since object defects occur at the pixel level, prompt learners often struggle to accurately determine whether an anomaly is present, limiting their effectiveness in fine-grained AD.

In this work, we propose a multi-class AD framework that puts together the strong descriptive power of VLMs and diffusion models, dubbed as {\ours}. It leverages textual descriptions extracted from off-the-shelf VLMs without requiring prompt learning. We devise one simple prompt to get the detailed description of the main object for the training. 
These descriptions are embedded using a text encoder. The encoded vector is then incorporated as an additional conditioning signal for the diffusion model to improve the reconstruction performance. Our approach exploits Latent Diffusion Models (LDMs)~\cite{ldm} for an efficient reconstruction process. 
Figure~\ref{fig:teaser} presents a high-level comparison between existing diffusion-based methods and {\ours}.

{\ours} is not limited to a specific domain, it shows superior performance on industrial (Real-IAD \cite{wang2024real}) and more general purpose (COCO-AD \cite{coco_ad}) large scale AD datasets. 
It significantly outperforms the state-of-the-art diffusion-based methods on pixel-level metrics. Notably, our method achieves up to a 25-point improvement on the Real-IAD~\cite{wang2024real} dataset in the pixel-level $PRO$ metric.
Unlike the existing diffusion-based methods~\cite{Fucka_2024_ECCV, zhang2023unsupervised, zhang2023diffusionad}, {\ours} supports multi-class AD where only a single model per dataset is trained rather than training a single model per class which would improve 1) \textbf{generalization} thanks to broader normality representation, 2) \textbf{robustness} thanks to reduced overfitting by avoiding per class training, 3) \textbf{scalability} thanks to reduced model complexity by maintaining and deploying a single model only, 4) \textbf{efficiency} thanks to faster training and inference, 5) \textbf{AD in complex scenarios} thanks to capturing contextual relationships between different object categories within a dataset, and 6) \textbf{detection of novel anomalies} thanks to the holistic representation of normality that is not constrained to any specific class-based definition.
The proposed {\ours} has the following key contributions:
\begin{itemize}
    \item We leverage the strong descriptive capabilities of vision-language models (VLMs) to guide the learning of the diffusion-based multi-class AD model.
    \item Without requiring specially designed or learned prompts, {\ours} substantially improves the baseline pixel-level performance.
    \item Learning a multi-class AD model rather than an individual class per category. This has a tangible impact on generalization, scalability and efficiency.
    \item We demonstrate the effectiveness of our approach on two diverse datasets, highlighting its robustness and generalizability.
\end{itemize}

\section{Related Work}
\label{sec:rel_work}

In this section, we first give a summary of prominent visual AD methods, then focus on the diffusion-based approaches, and finally elaborate on the prompt learning based methods which use textual descriptions as an additional modality for AD.

\noindent\textbf{Visual AD.}
Existing methods can be broadly categorized into three strategies: embedding-based, augmentation-based, and reconstruction-based.

Embedding based approaches~\cite{s18010209, lee2022cfa, gudovskiy2022cflow, roth2022towards, lei2023pyramidflow} first extract high-dimensional image or patch features from pretrained networks, and then compare features coming from normal and anomalous data to locate anomalies. Very first approaches~\cite{s18010209} use clustering on the extracted patch features from ResNet~\cite{resnet} pretrained on ImageNet~\cite{deng2009imagenet}. Later works~\cite{spade, padim, roth2022towards, lee2022cfa} add memory banks to extend the available feature context. Normalizing flow based methods~\cite{gudovskiy2022cflow, lei2023pyramidflow} extract multi-level patch features and learn either the distribution of positive patches or minimize the distance between each positive patch. Knowledge distillation is also used within a teacher-student framework in~\cite{Bergmann_Fauser_Sattlegger_Steger_2020, Wang_Wu_Cui_Shen_2021}. 

Another line of work~\cite{zavrtanik2021draem, liu2023simplenet, zhang2024realnet,FAUAD} employs augmentations to simulate anomalous data. DRAEM~\cite{zavrtanik2021draem} trains an autoencoder to reconstruct both normal and generated noisy images using Perlin~\cite{perlin}, and a discriminative network locates the anomalies. 

Generative methods learn to reconstruct normal samples using VAE~\cite{bergmann2018improving}, GAN~\cite{akcay2019ganomaly, schlegl2019f}, ViT~\cite{zhang2023exploring}, or more recently, diffusion models~\cite{wolleb2022diffusion,wyatt2022anoddpm,zhang2023diffusionad, he2024diffusion}. 
During inference, a higher reconstruction error will result from anomalous inputs compared to the normal ones.
Bergmann et al.~\cite{bergmann2018improving} show a VAE trained with a structural
similarity (SSIM) objective that is able to detect pixel-level defects. GANomaly~\cite{akcay2019ganomaly} employs a GAN architecture to detect image-level anomalies. ViTAD~\cite{zhang2023exploring} uses vision transformers in VAE architecture and obtains state-of-the-art performance.

\noindent\textbf{Diffusion Models.} 
These models~\cite{ddpm, ldm} show unprecedented image generation capabilities, which makes them a natural candidate for AD. The first works using diffusion are proposed in the medical domain~\cite{wolleb2022diffusion, wyatt2022anoddpm}. Later works~\cite{zhang2023unsupervised, zhang2023diffusionad, he2024diffusion, livernochediffusion, Fucka_2024_ECCV, yao2024glad} extend diffusion models to industrial defect localization. DiffAD~\cite{zhang2023unsupervised} consists of two sub-networks: a Latent Diffusion Model (LDM) to reconstruct samples and a discriminative network to segment defect locations. DiffusionAD~\cite{zhang2023diffusionad} follows a similar approach as DiffAD with the addition of a synthetic anomaly generation pipeline. During training, synthetic anomaly data is generated either using an external dataset~\cite{cimpoi2014describing} for overlapping classes or Perlin noise~\cite{perlin} similar to DRAEM~\cite{zavrtanik2021draem}. In order to reduce the inference time, a norm-guided one-step denoising approach is employed. The main drawbacks of these approaches are that, first, they train separate models for each class, which limits them from generalizing to large datasets with many objects, and second, the generation and segmentation models are separately trained. Transfusion~\cite{Fucka_2024_ECCV} merges the generation and anomaly localization parts into a single diffusion model which works in image space~\cite{ddpm}. The diffusion model takes input as Perlin noise added to normal images along with noise masks and outputs both the noise-removed images along with the noise masks.  

Recently, DiAD~\cite{he2024diffusion} proposed using a conditional diffusion model which relaxed the need to train separate models for each class. Moreover, the anomaly segmentation is done by comparing the multi-level features extracted from a fixed pretrained network for input and reconstructed images, instead of training a separate segmentation network. These two properties make DiAD versatile to adapt to different datasets which could have multiple objects in the images as anomalies. However, the performance of DiAD and previous diffusion-based methods is falling behind the current state-of-the-art on large and complex datasets~\cite{wang2024real, coco_ad}, most likely due to the weak guidance they are using. Our method strengthens the guidance by using detailed text descriptions extracted from powerful VLMs.

\noindent\textbf{Multi Modal Methods.}
Alternatively, so far, text data is used in the context of prompt learning for extracting short anomaly descriptions. These methods utilize CLIP~\cite{clip} which shows exceptional zero-shot detection capability in various unsupervised vision tasks. CLIP-AD~\cite{liznerski2022exposing} is one of the very first approaches using plain CLIP image and text features for image-level AD. Later methods~\cite{winclip, zhouanomalyclip, gu2024filo, cao2025adaclip} focus on either crafting or learning useful text prompts that capture the possible states of anomaly classes. However, in order to learn useful prompts, these methods utilize external anomaly data. Different from these, LAVAD~\cite{zanella2024harnessing} proposes a training-free video AD framework supported by VLMs. They generate detailed captions using Blip-2~\cite{li2023blip} for each frame in a video sequence and measure the alignment between CLIP image and text encoders. We also follow a prompt learning-free methodology in our work. A recent study by Xu et al.~\cite{xu2025towards} introduces a novel human-annotated dataset for anomaly detection enriched with detailed anomaly descriptions, and shows that fine-tuning VLMs on this dataset significantly improves both detection performance and reasoning ability. However, this method lacks anomaly localization. For a more comprehensive review of VLMs in AD, we refer to the following surveys~\cite{miyai2024generalized, xu2024large}.
\begin{figure*}[t]
\centering
\includegraphics[width=1.0\linewidth]{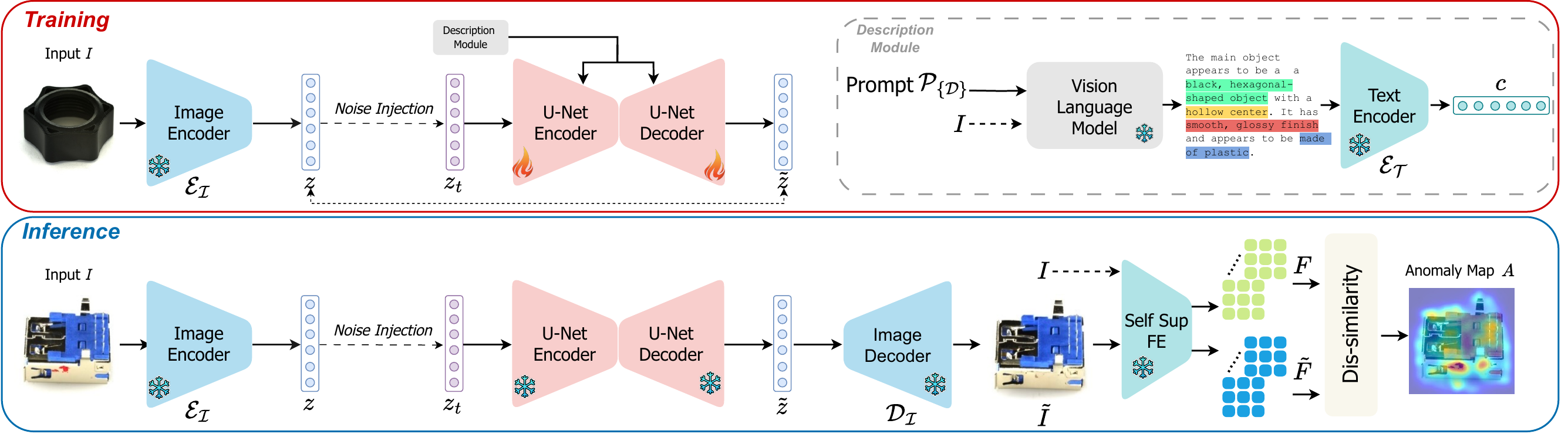}
\caption{The processing pipeline of {\ours}. During training (top), a normal (i.e. anomaly-free) image is fed to both 1) an off-the-shelf VLM (on the right) to extract the detailed description of the object using the query ($\mathcal{P}_D$), which is further encoded into condition vector $c$ using text encoder, and to 2) a pretrained image encoder (note that we finetune the image autoencoder using only normal images beforehand) to get the latent vector. Then, a diffusion process adds noise to the latent vector and learns to denoise it with the guidance coming from the condition vector. During inference (bottom), the same process as training is followed, except there is no text description is used to condition the diffusion model. The denoised latent code is fed to the pretrained image decoder to get the reconstructed image. Anomaly segmentation is done by finding the dissimilar locations on the feature maps of input and reconstructed normal images.} 
\label{fig:model}
\vspace{-3mm}
\end{figure*}

\section{{\ours}}
\label{sec:method}

We present the detailed structure of our method in Figure~\ref{fig:model}. The training pipeline starts by feeding the input image $I$ to the pretrained image encoder $\mathcal{E}_I$ to extract the latent representation $z$. In parallel, we forward the image to an off-the-shelf large VLM along with a generic prompt $\mathcal{P}$ to get the image description VLM$(I, \mathcal{P})$. The image description is further encoded into a condition vector $c$ using a pretrained text encoder $\mathcal{E}_T$. The diffusion process starts by adding noise $T$ times to the latent vector $z$ and getting the noised latent vector $z_t$. Then, denoising network recovers input $z$ from noisy latent $z_t$ with the guidance coming from the conditional vector $c$. Note that, unlike the current state-of-the-art diffusion-based methods, which train separate models per category, we train a single model per dataset.

At inference time, the denoised latent $\tilde{z}$ is decoded back to the image space using the pretrained image decoder $\mathcal{D}_I$ to get the reconstructed image $\tilde{I}$. In order to segment the anomaly locations, the input image $I$ and the reconstructed image $\tilde{I}$ are fed to an off-the-shelf self-supervised model to extract the features $F$ and $\tilde{F}$, respectively. Finally, the pixels with high dissimilarity between these two feature vectors denote the Anomaly Map $A$.

\noindent\textbf{Vision Language Model (VLM)} The mainstream approach to guide the unsupervised diffusion-based AD methods is to use some form of noisy input generation by either using external defect datasets or generating synthetic defects using noise generators. We argue that this form of guidance is both unrealistic for specific anomaly types like scratches and does not generalize to natural image datasets such as COCO-AD~\cite{coco_ad}. Instead, we propose to guide the diffusion model with detailed image descriptions using large VLMs.
Considering the main goal of the training pipeline is to learn to reconstruct the input latent perfectly, we extract detailed descriptions of the input images that could help this objective.

For industrial datasets (e.g. Real-IAD \cite{wang2024real}), the normal training images are examined by the VLM model by a prompt to describe the main object. The description query is $\mathcal{P}_D$: "\textit{Describe the main object in detail.}". 
Whereas for the inference part, that has defective objects as well, we observe that prompting the VLM to describe the object results in noisy descriptions. We decide to use no text description for the inference part and let the diffusion model generate unconditionally.  
Hence the prompt used for training becomes $\mathcal{P}_T = \mathcal{P}_D$, and during inference we do not use any text conditioning.

For natural image datasets (e.g. COCO-AD \cite{coco_ad}), the anomalies are object-level (i.e. unseen classes during training) not pixel-level, and VLMs show exceptional performance to describe the objects, even segment them~\cite{seem, seed}, in an image as they are trained in this manner. Thus, we use the description query with a slight modification for both training and inference (i.e. $\mathcal{P}_T = \mathcal{P}_I= \mathcal{P}_D$). Our description query $\mathcal{P}_D$ on COCO-AD is: ``\textit{Describe the visual features of image in detail.}". In this way, we expect the VLM to mention the objects and their relations, which will guide the diffusion model. These descriptions are extracted without using any label information. Finally, the description is encoded using a text encoder (e.g. CLIP~\cite{clip}) into the text condition vector $c$. 
For a detailed analysis on the effect of using queries, refer to Section Ablation Experiments.

\noindent\textbf{Diffusion Model} We use the same diffusion architecture presented in~\cite{ldm} which comprises an image autoencoder and a denoising attention-based U-Net~\cite{ronneberger2015u} model. In order to extract the latent vectors from images, we first train the image autoencoder using only the training set which contains normal images.  

For a given input image $I \in \mathbb{R}^{ H\times W\times 3}$ in pixel space, we first encode it into latent space $z = \mathcal{E_I}(I)$ where $z \in \mathbb{R}^{h\times w\times d}$ using the trained image encoder $\mathcal{E_I}$. The forward process can be characterized by: 
\begin{equation}
z_t=\sqrt{\bar{\alpha}_t} z+\sqrt{1-\bar{\alpha}_t} \epsilon_t, \epsilon_t \sim \mathcal{N}(\mathbf{0}, \mathbf{I}).
\end{equation}

where \( \alpha_t = 1 - \beta_t \), and the cumulative product \( \bar{\alpha}_t \) is given by \( \bar{\alpha}_t = \prod_{i=1}^{T} \alpha_i = \prod_{i=1}^{T} (1 - \beta_i) \). Here, \( \beta_i \) defines the noise schedule, which controls the amount of noise introduced at each timestep.

The noised representation $z_t$, and the text condition vector $c$ are fed into both the denoising U-Net encoder and decoder networks. After $T$ steps of the reverse denoising process, the denoised latent vector $\tilde{z}$ is obtained. In the end, the reconstructed image $\tilde{I}$ is generated by forwarding the $\tilde{z}$ to the pretrained image decoder $\mathcal{D_I}(\tilde{z})$.
The training objective of {\ours} is:
\begin{equation}
\mathcal{L}_{{\ours}}=\mathbb{E}_{z, t, c, \epsilon \sim \mathcal{N}(0,1)}\left[\left\|\epsilon-\epsilon_\theta\left(z_t, t, c\right)\right\|_2^2\right].
\end{equation}
where the objective function aims to minimize the difference between the noise added to the data at a certain time step $\epsilon$ and the noise $\epsilon_\theta$ predicted by the model.
Note that our diffusion model is trained in a multi-class setting by using all the normal images from training sets without utilizing any class labels.

\renewcommand{\arraystretch}{1}
\begin{figure*}
\captionsetup[subfigure]{labelformat=empty}
\centering
\resizebox{1.0\textwidth}{!}{
\begin{tabular}{l@{}c@{}c@{}c@{}c@{}c@{}c@{}c@{}c}
& \tiny{phone battery} & \tiny{plastic nut} & \tiny{sim card} & \tiny{toy brick} & \tiny{u block} & \tiny{wooden beads} & \tiny{woodstick}& \tiny{zipper}  \\

{\rotatebox[origin=t]{90}{\textit{{I}}}}  & 
\includegraphics[width=\myww,  ,valign=m, keepaspectratio,] {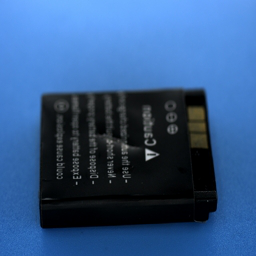} &
\includegraphics[width=\myww,  ,valign=m, keepaspectratio,] {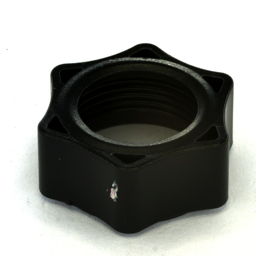} & 
\includegraphics[width=\myww,  ,valign=m, keepaspectratio,] {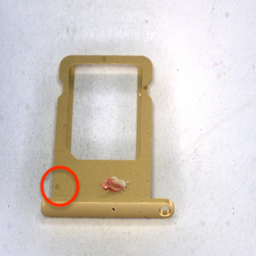}  &
\includegraphics[width=\myww,  ,valign=m, keepaspectratio,] {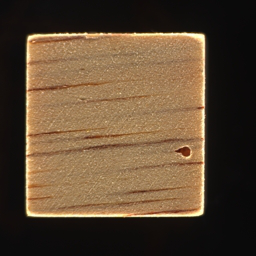} & 
\includegraphics[width=\myww,  ,valign=m, keepaspectratio,] {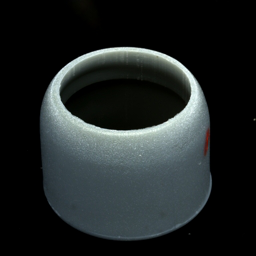}  &
\includegraphics[width=\myww,  ,valign=m, keepaspectratio,] {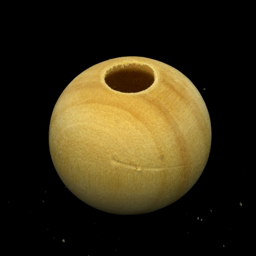} & 
\includegraphics[width=\myww,  ,valign=m, keepaspectratio,] {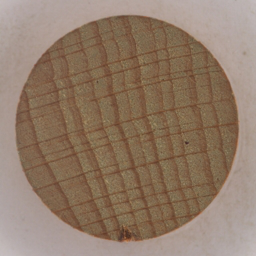}  &
\includegraphics[width=\myww,  ,valign=m, keepaspectratio,] {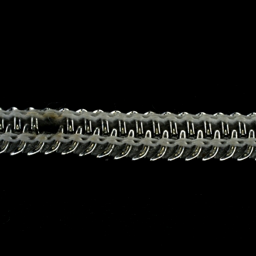} \\

{\rotatebox[origin=t]{90}{\textit{\textbf{ \tiny Anomaly Map}}}}&  

\includegraphics[width=\myww,  ,valign=m, keepaspectratio,] {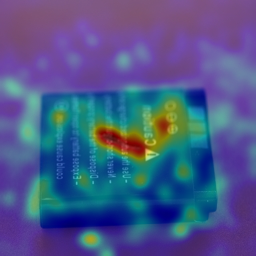}  &
\includegraphics[width=\myww,  ,valign=m, keepaspectratio,] {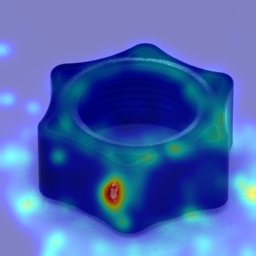} &  
\includegraphics[width=\myww,  ,valign=m, keepaspectratio,] {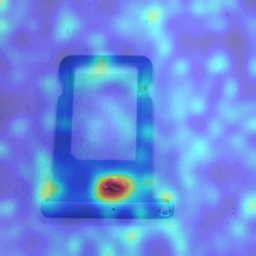}  &
 \includegraphics[width=\myww,  ,valign=m, keepaspectratio,] {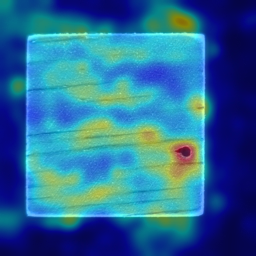} &
\includegraphics[width=\myww,  ,valign=m, keepaspectratio,] {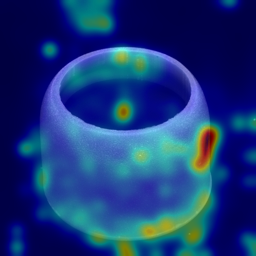}  &
\includegraphics[width=\myww,  ,valign=m, keepaspectratio,] {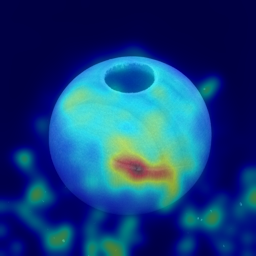} &
\includegraphics[width=\myww,  ,valign=m, keepaspectratio,] {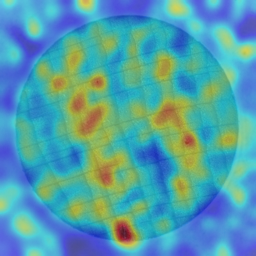}  &
\includegraphics[width=\myww,  ,valign=m, keepaspectratio,] {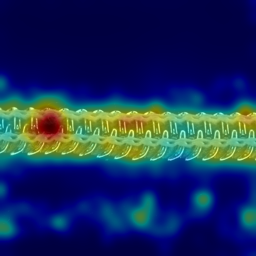} \\

{\rotatebox[origin=t]{90}{\textit{\textbf{\tiny Mask}}}} & 
\includegraphics[width=\myww,  ,valign=m, keepaspectratio,] {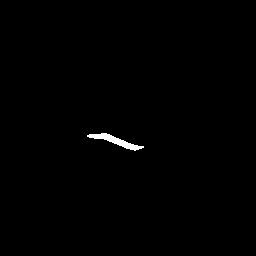}  &
\includegraphics[width=\myww,  ,valign=m, keepaspectratio,] {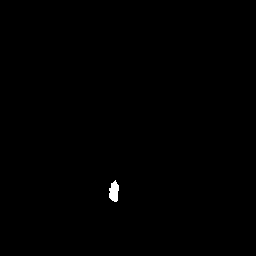} &
\includegraphics[width=\myww,  ,valign=m, keepaspectratio,] {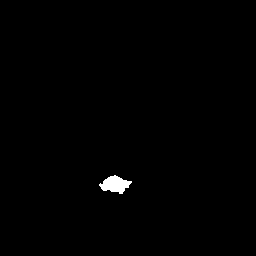}  &
\includegraphics[width=\myww,  ,valign=m, keepaspectratio,] {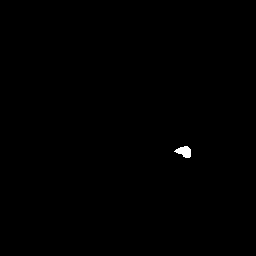} &
\includegraphics[width=\myww,  ,valign=m, keepaspectratio,] {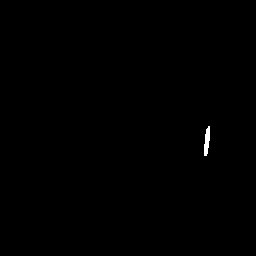}  &
\includegraphics[width=\myww,  ,valign=m, keepaspectratio,] {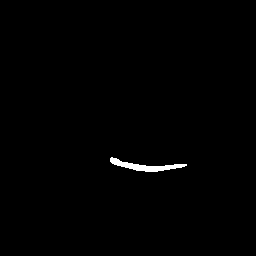}  &
\includegraphics[width=\myww,  ,valign=m, keepaspectratio,] {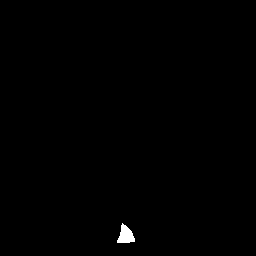}  &
\includegraphics[width=\myww,  ,valign=m, keepaspectratio,] {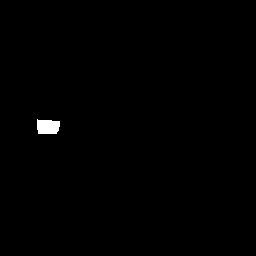} \\

\end{tabular}}
\caption{Example anomalous images from Real-IAD dataset, and predicted anomaly segmentation maps using {\ours}.}
\label{fig:vlm_desc}
\vspace{-3mm}
\end{figure*}

\noindent\textbf{Anomaly Segmentation Model} 
The anomaly segmentation part is only used during inference to localize the defective pixels. 
For a given test image $I$, after following the same process as training, the image decoder $\mathcal{D_I}$ reconstructs the denoised latent vector $\tilde{z}$ to image space $\tilde{I}$. The difference between the input and reconstructed image will denote the defective pixels.    

The input image $I$ and the reconstructed image $\tilde{I}$ are fed to a pretrained model to extract the features $F$ and $\tilde{F}$, respectively. Then we resize these patch features to the input image size and calculate the pixel-wise cosine similarity. Finally, we calculate $1- cossim( \tilde{F}, F)$ to localize the anomaly regions. In order to maintain the unsupervised nature of our model, we utilize a self-supervised method DINO~\cite{dino} for the anomaly segmentation, and use the patch features from the last layer.

\section{Experiments}
\label{sec:exp}

\subsection{Datasets}
\label{sec:datasets}

We experimented with two very large datasets, industrial defect detection Real-IAD~\cite{wang2024real}, and real-world anomaly detection COCO-AD~\cite{coco_ad}. 
Real-IAD contains $30$ objects with multiple images from different views and has orders of magnitude more training and test images compared to previous defect detection datasets like MVTec-AD~\cite{bergmann2019mvtec} and VISA~\cite{visa}. Moreover, the dataset includes a diverse range of defect types, including very small and challenging anomalies. 

COCO-AD~\cite{coco_ad} fills the need for having a real-world anomaly localization dataset. It is derived from the original COCO~\cite{coco} dataset by choosing consecutive 20 classes as anomalies out of present $80$ classes in each split. The images that contain any one of the selected $20$ classes are removed from the training sets. Similarly, validation set is constructed by selecting the images which contain any anomaly classes labeled as anomalies, and the rest are labeled as normal.
Details of both datasets are presented in Table~\ref{tab:dataset}.

\begin{table}
    \centering
    \setlength\tabcolsep{4.0pt}
    \resizebox{1.0\linewidth}{!}{
        \begin{tabular}{cccccccc}
        \toprule
        \multirow{3}{*}{Dataset} & \multicolumn{2}{c}{Categories} & & \multicolumn{4}{c}{Images}  \\
        \cline{2-3} \cline{5-8} 
        & \multirow{2}{*}{Train} & \multirow{2}{*}{Test} & & Train & & \multicolumn{2}{c}{Test} \\
        \cline{5-5} \cline{7-8}
        & & & & Normal & & Anomaly & Normal \\
        \toprule
        Real-IAD~\cite{wang2024real}      & 30 & 	30 & & 36,465 & & 51,329& 63,256  \\
        \hline
        \multirow{4}{*}{COCO-AD~\cite{coco_ad}} & \multirow{4}{*}{61} & \multirow{4}{*}{81} & & 30,438 & & 1,291& 3,661  \\
        & & & & 65,133 & & 2,785& 2,167 \\
        & & & & 79,083 & & 3,328& 1,624 \\
        & & & & 77,580 & & 3,253& 1,699 \\
        \bottomrule
        \end{tabular}}
    \caption{Details of the datasets used in experiments.}
    \label{tab:dataset}
    \vspace{-3mm}
\end{table}

\renewcommand{\arraystretch}{1}
\begin{figure*}
\captionsetup[subfigure]{labelformat=empty}
\centering
\resizebox{1.0\textwidth}{!}{
\begin{tabular}{l@{}c@{}c@{}c@{}c@{}c@{}c@{}c@{}c@{}c@{}c}
& \tiny{button battery} & \tiny{end cap} & \tiny{eraser} & \tiny{fire hood} & \tiny{mint} & \tiny{mounts} & \tiny{pcb}& \tiny{transistor} & \tiny{woodstick} & \tiny{zipper} \\

{\rotatebox[origin=t]{90}{\textit{{I}}}}  & 
\includegraphics[width=\mywidth,  ,valign=m, keepaspectratio,] {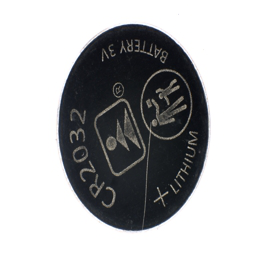} &
\includegraphics[width=\mywidth,  ,valign=m, keepaspectratio,] {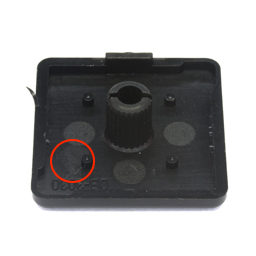} &
\includegraphics[width=\mywidth,  ,valign=m, keepaspectratio,] {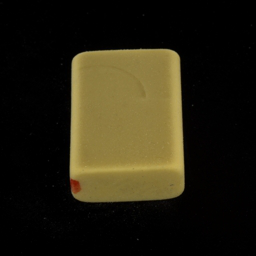} &
\includegraphics[width=\mywidth,  ,valign=m, keepaspectratio,] {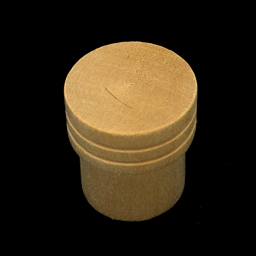} &
\includegraphics[width=\mywidth,  ,valign=m, keepaspectratio,] {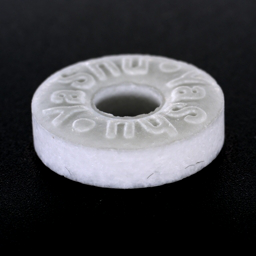} &
\includegraphics[width=\mywidth,  ,valign=m, keepaspectratio,] {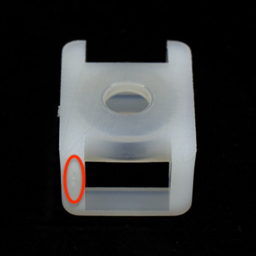} &
\includegraphics[width=\mywidth,  ,valign=m, keepaspectratio,] {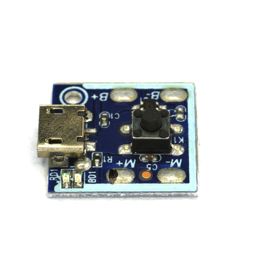} &
\includegraphics[width=\mywidth,  ,valign=m, keepaspectratio,] {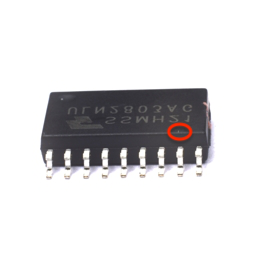}  &
\includegraphics[width=\mywidth,  ,valign=m, keepaspectratio,] {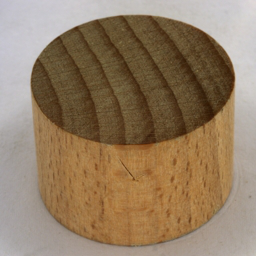} &
\includegraphics[width=\mywidth,  ,valign=m, keepaspectratio,] {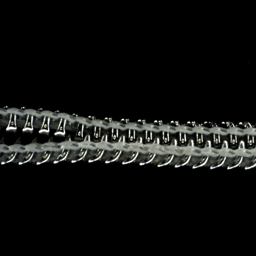} \\

{\rotatebox[origin=t]{90}{\textit{\textbf{ \tiny DiAD}}}}&  

\includegraphics[width=\mywidth,  ,valign=m, keepaspectratio,] {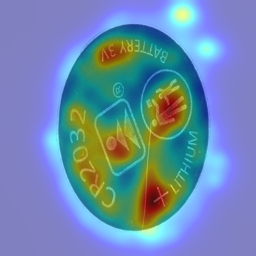} &
\includegraphics[width=\mywidth,  ,valign=m, keepaspectratio,] {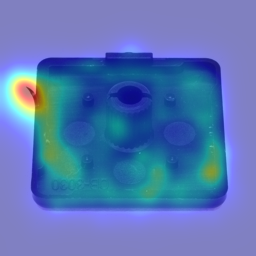} &
\includegraphics[width=\mywidth,  ,valign=m, keepaspectratio,] {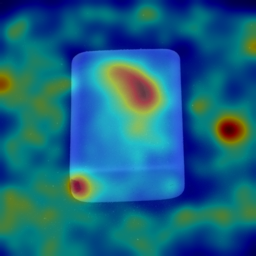} &
\includegraphics[width=\mywidth,  ,valign=m, keepaspectratio,] {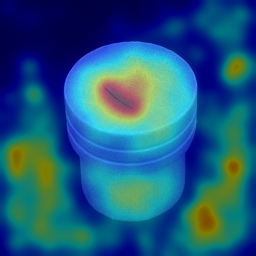} &
\includegraphics[width=\mywidth,  ,valign=m, keepaspectratio,] {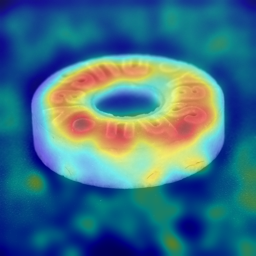} &
\includegraphics[width=\mywidth,  ,valign=m, keepaspectratio,] {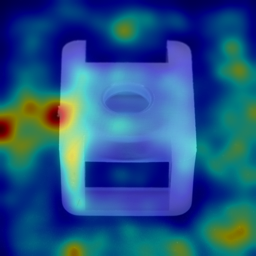} &
\includegraphics[width=\mywidth,  ,valign=m, keepaspectratio,] {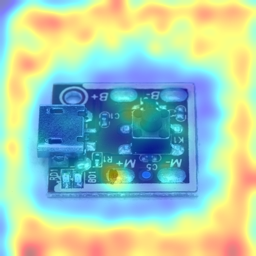} &
\includegraphics[width=\mywidth,  ,valign=m, keepaspectratio,] {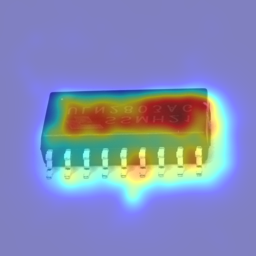}  &
\includegraphics[width=\mywidth,  ,valign=m, keepaspectratio,] {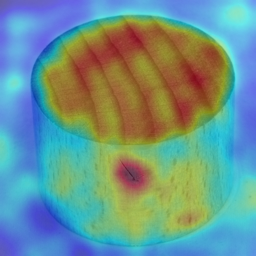} &
\includegraphics[width=\mywidth,  ,valign=m, keepaspectratio,] {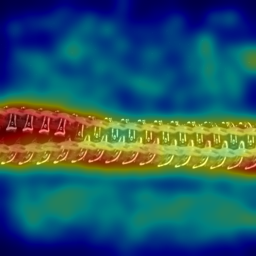} \\

{\rotatebox[origin=t]{90}{\textit{\textbf{\tiny TransFusion}}}} & 
\includegraphics[width=\mywidth,  ,valign=m, keepaspectratio,] {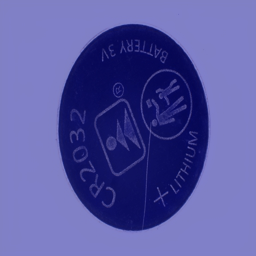} &
\includegraphics[width=\mywidth,  ,valign=m, keepaspectratio,] {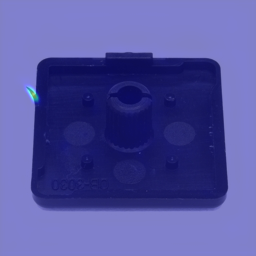} &
\includegraphics[width=\mywidth,  ,valign=m, keepaspectratio,] {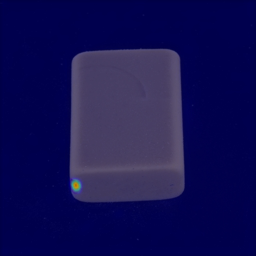} &
\includegraphics[width=\mywidth,  ,valign=m, keepaspectratio,] {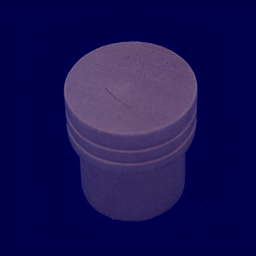} &
\includegraphics[width=\mywidth,  ,valign=m, keepaspectratio,] {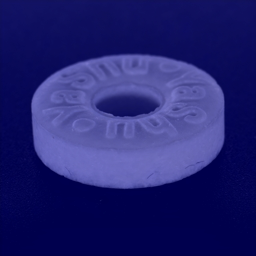} &
\includegraphics[width=\mywidth,  ,valign=m, keepaspectratio,] {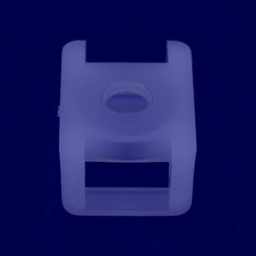} &
\includegraphics[width=\mywidth,  ,valign=m, keepaspectratio,] {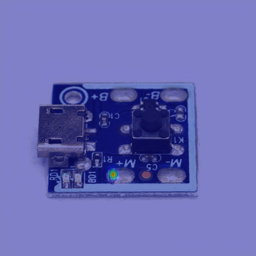} &
\includegraphics[width=\mywidth,  ,valign=m, keepaspectratio,] {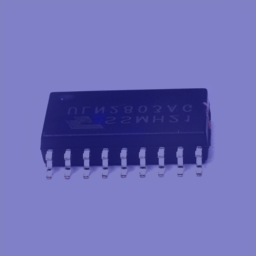}  &
\includegraphics[width=\mywidth,  ,valign=m, keepaspectratio,] {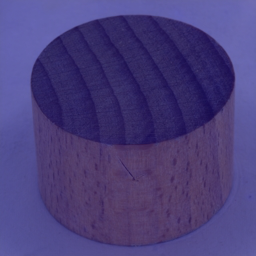} &
\includegraphics[width=\mywidth,  ,valign=m, keepaspectratio,] {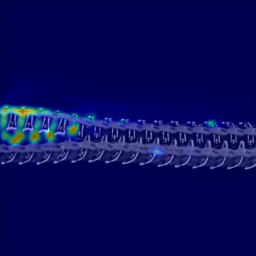} \\

{\rotatebox[origin=t]{90}{\textit{\textbf{\tiny {\ours}}}}} & 
\includegraphics[width=\mywidth,  ,valign=m, keepaspectratio,] {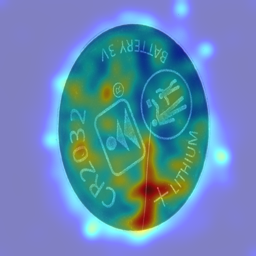} &
\includegraphics[width=\mywidth,  ,valign=m, keepaspectratio,] {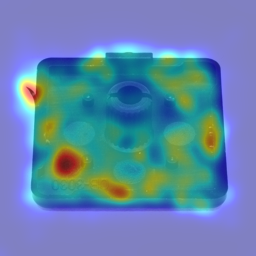} &
\includegraphics[width=\mywidth,  ,valign=m, keepaspectratio,] {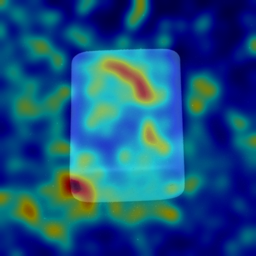} &
\includegraphics[width=\mywidth,  ,valign=m, keepaspectratio,] {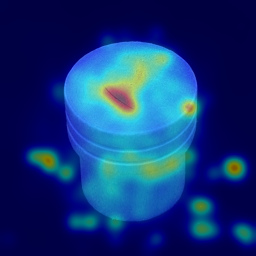} &
\includegraphics[width=\mywidth,  ,valign=m, keepaspectratio,] {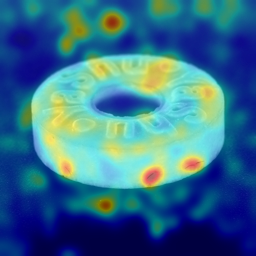} &
\includegraphics[width=\mywidth,  ,valign=m, keepaspectratio,] {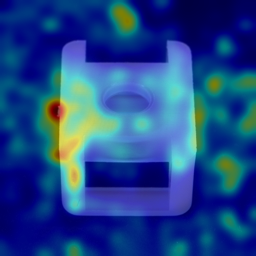} &
\includegraphics[width=\mywidth,  ,valign=m, keepaspectratio,] {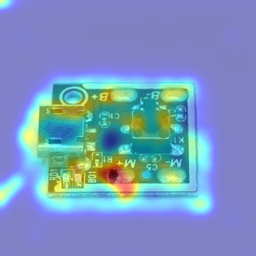} &
\includegraphics[width=\mywidth,  ,valign=m, keepaspectratio,] {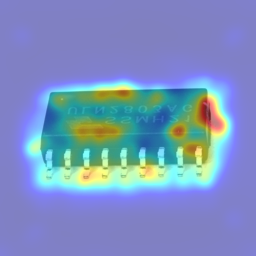}  &
\includegraphics[width=\mywidth,  ,valign=m, keepaspectratio,] {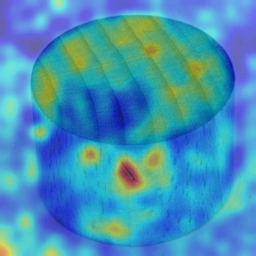} &
\includegraphics[width=\mywidth,  ,valign=m, keepaspectratio,] {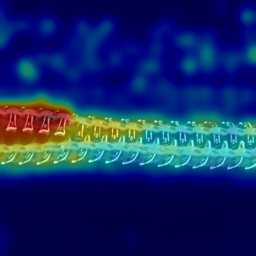} \\

{\rotatebox[origin=t]{90}{\textit{\textbf{\tiny Mask}}}} & 
\includegraphics[width=\mywidth,  ,valign=m, keepaspectratio,] {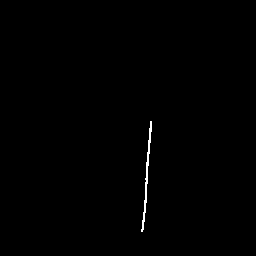} &
\includegraphics[width=\mywidth,  ,valign=m, keepaspectratio,] {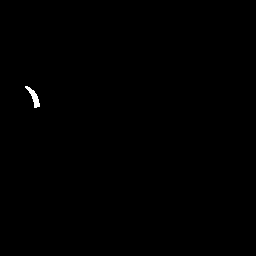} &
\includegraphics[width=\mywidth,  ,valign=m, keepaspectratio,] {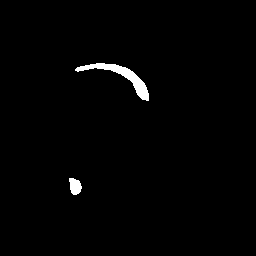} &
\includegraphics[width=\mywidth,  ,valign=m, keepaspectratio,] {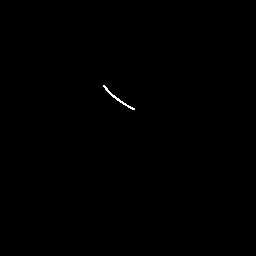} &
\includegraphics[width=\mywidth,  ,valign=m, keepaspectratio,] {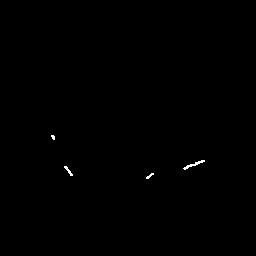} &
\includegraphics[width=\mywidth,  ,valign=m, keepaspectratio,] {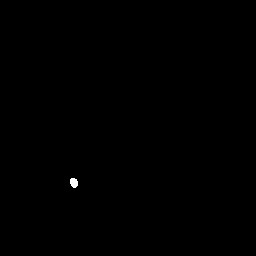} &
\includegraphics[width=\mywidth,  ,valign=m, keepaspectratio,] {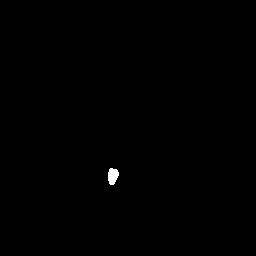} &
\includegraphics[width=\mywidth,  ,valign=m, keepaspectratio,] {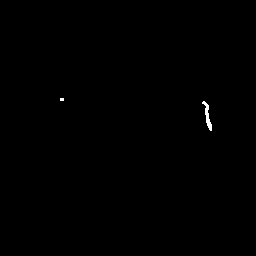}  &
\includegraphics[width=\mywidth,  ,valign=m, keepaspectratio,] {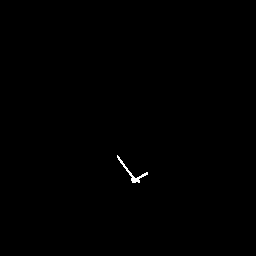} &
\includegraphics[width=\mywidth,  ,valign=m, keepaspectratio,] {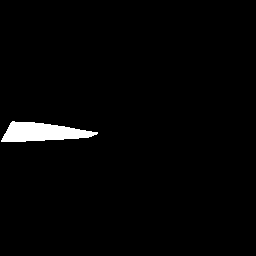} \\

\end{tabular}}
\caption{Visual comparison of diffusion-based methods on Real-IAD dataset. DiAD detects the anomaly locations with the expense of having many false positives.}
\vspace{-5mm}
\label{fig:realiad_comparison}
\end{figure*}

\subsection{Experimental Setup}
\label{sec:setup}

We implemented our method using PyTorch~\cite{pytorch}. For all datasets, images are resized to $256\times256$ pixels before training. All models are trained on Nvidia A100 GPUs.
Below, we give implementation and training details for all sub-networks.  

\noindent\textbf{Image Autoencoder} is trained using only normal training (anomaly-free) images with KL divergence loss on the latent space, and GAN~\cite{gan} objective on the image space. We trained the autoencoder for $50$ epochs on Real-IAD and COCO-AD datasets using a batch size of $32$ and a learning rate of $4.5e-5$. Our autoencoder implementation is the same as the one used in LDM~\cite{ldm}.

\noindent\textbf{Denoising Model} 
Denoising U-Net weights are initialized using the pretrained Stable Diffusion v1.5 weights~\footnote{\raggedright huggingface.co/stable-diffusion-v1-5/stable-diffusion-v1-5\par}. We trained our models with a batch size of $12$ and a learning rate of $1e-5$ using the training approach presented in ControlNet~\cite{zhang2023adding}. LDM is conditioned on CLIP-encoded~\cite{clip} text descriptions extracted from VLM.  

\noindent\textbf{VLM.} Text descriptions for train  images are extracted using InternVL-2~\cite{captioner} $8B$ version which achieves state-of-the-art performance on various benchmarks. We did not conduct any prompt learning or searching to devise $\mathcal{P}_D$.

\noindent\textbf{State-of-the-art comparison.} {\ours} is compared with methods from all AD categories and the results are reported in Tables~\ref{tab:real_iad_scores} and \ref{tab:coco_ad_scores}. The most recent diffusion-based multi-class (DiAD~\cite{he2024diffusion}) and single-class (TransFusion~\cite{Fucka_2024_ECCV}) methods are included. All methods are trained for $100$ epochs using their official codes. To compare with other methods, we used the results reported in a recent survey~\cite{ader} from the authors of the Real-IAD \cite{wang2024real} and COCO-AD \cite{coco_ad} datasets. 

In the supplementary material, we present per-class results on Real-IAD, per-split results on COCO-AD, additional quantitative results on the MVTec-AD~\cite{bergmann2019mvtec} and VISA~\cite{visa} datasets, and more visual comparisons on Real-IAD and COCO-AD datasets.


\subsection{Qualitative Results}
\label{sec:visual_res}

Anomaly maps of {\ours} are shown in Figure~\ref{fig:vlm_desc}, and  a visual comparison of diffusion-based methods on a subset of Real-IAD classes is presented in Figure~\ref{fig:realiad_comparison}. DiAD detects many normal locations as anomalies with high confidence. Whereas Transfusion misses many anomalous pixels, and the detected ones have low confidence values. Especially when the defects are not very severe, such as scratches, TransFusion could not localize, most likely due to the employed noise generation scheme. 
{\ours} detects various types of defects correctly, and even sometimes locates missed annotations (denoted as red circles) as in the cases of \textit{end cap}, \textit{mounts} and \textit{transistor} classes. 
This focus on fine-grained localization, however, is leading to a lower $ROC_I$ score compared to baselines.

\begin{figure}
\captionsetup[subfigure]{labelformat=empty}
\centering
\setlength\tabcolsep{1.5pt} 
\resizebox{1.0\columnwidth}{!}{
\begin{tabular}{l@{}m{0.2\columnwidth}m{0.2\columnwidth}m{0.2\columnwidth}m{0.2\columnwidth}}
& \tiny{Split-0} & \tiny{Split-1} & \tiny{Split-2} & \tiny{Split-3}  \\

{\rotatebox[origin=t]{90}{\textit{{\tiny I}}}}  & 
\includegraphics[width=\mywidth,  ,valign=m, keepaspectratio,] {./figures/vis_comp/coco/part0/000000000872.jpg_true.png}  &
\includegraphics[width=\mywidth,  ,valign=m, keepaspectratio,] {./figures/vis_comp/coco/part1/000000003553.jpg_true.png} &
\includegraphics[width=\mywidth,  ,valign=m, keepaspectratio,] {./figures/vis_comp/coco/part2/000000077595.jpg_true.png} &
\includegraphics[width=\mywidth,  ,valign=m, keepaspectratio,] {./figures/vis_comp/coco/part3/000000001503.jpg_true.png} \\

{\rotatebox[origin=t]{90}{\textit{\textbf{ \tiny DiAD}}}} &  
\includegraphics[width=\mywidth,  ,valign=m, keepaspectratio,] {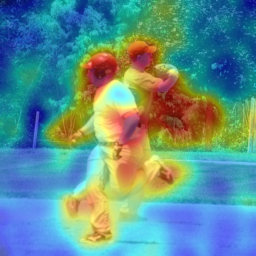}  &
\includegraphics[width=\mywidth,  ,valign=m, keepaspectratio,] {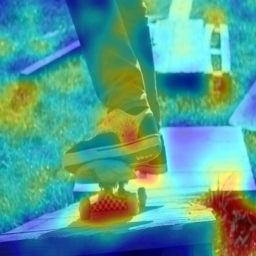} &
\includegraphics[width=\mywidth,  ,valign=m, keepaspectratio,] {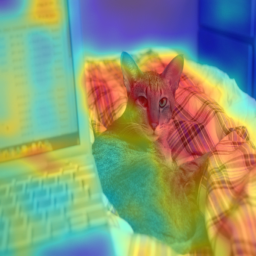} &
\includegraphics[width=\mywidth,  ,valign=m, keepaspectratio,] {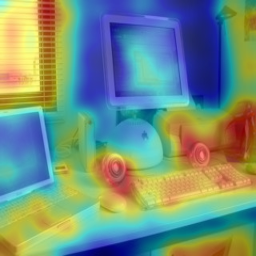} \\

{\rotatebox[origin=t]{90}{\textit{\textbf{\tiny TransFusion}}}} & 
\includegraphics[width=\mywidth,  ,valign=m, keepaspectratio,] {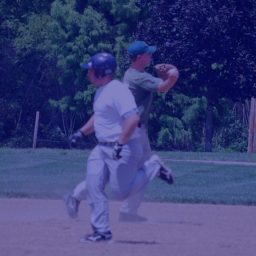}  &
\includegraphics[width=\mywidth,  ,valign=m, keepaspectratio,] {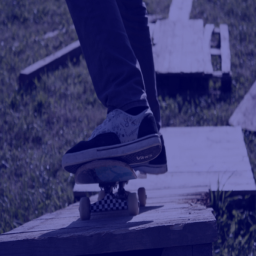} &
\includegraphics[width=\mywidth,  ,valign=m, keepaspectratio,] {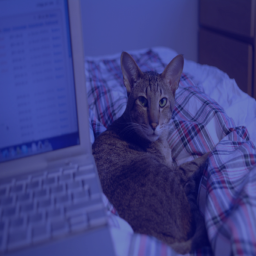} &
\includegraphics[width=\mywidth,  ,valign=m, keepaspectratio,] {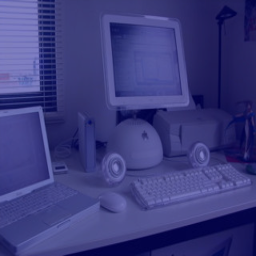} \\

{\rotatebox[origin=t]{90}{\textit{\textbf{\tiny {\ours}}}}} & 
\includegraphics[width=\mywidth,  ,valign=m, keepaspectratio,] {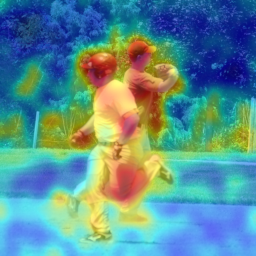}  &
\includegraphics[width=\mywidth,  ,valign=m, keepaspectratio,] {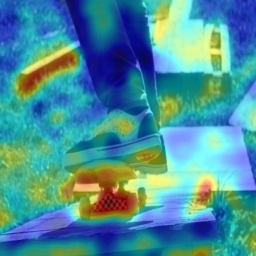} &
\includegraphics[width=\mywidth,  ,valign=m, keepaspectratio,] {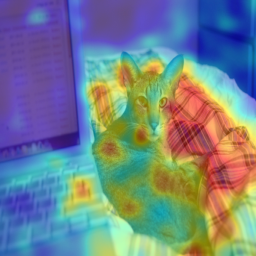} &
\includegraphics[width=\mywidth,  ,valign=m, keepaspectratio,] {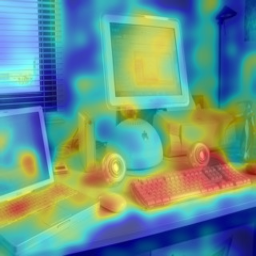} \\

{\rotatebox[origin=t]{90}{\textit{\textbf{\tiny Mask}}}} & 
\includegraphics[width=\mywidth,  ,valign=m, keepaspectratio,] {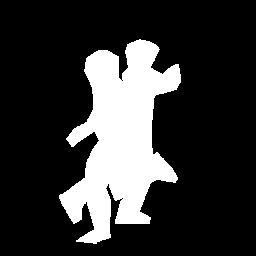}  &
\includegraphics[width=\mywidth,  ,valign=m, keepaspectratio,] {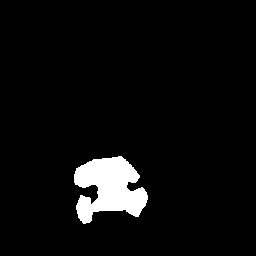} &
\includegraphics[width=\mywidth,  ,valign=m, keepaspectratio,] {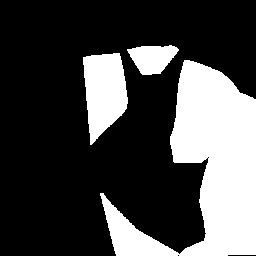} &
\includegraphics[width=\mywidth,  ,valign=m, keepaspectratio,] {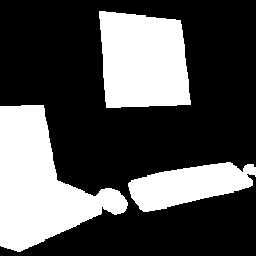} \\
\end{tabular}}
\caption{Visual comparison of diffusion-based methods on COCO-AD dataset.}
\vspace{-3mm}
\label{fig:cocoad_comparison}
\end{figure}

We show example results from four splits of the COCO-AD dataset in Figure~\ref{fig:cocoad_comparison}. TransFusion failed to detect many anomaly locations as expected, since the method is bound to a specific domain. On the other hand, the domain-independent methods, DiAD and {\ours}, successfully locate the anomalous objects even if the scene is highly complex. DiAD performs well when there is a single anomaly class present in a given image such as the first and second columns where \textit{person} and \textit{skateboard} are the only anomaly objects, respectively. However, in the last column where three anomaly classes are present, DiAD only detects parts of the \textit{keyboard} object correctly. 
On all splits, {\ours} achieves the best localization performance, even detecting both of the small \textit{mouse} objects in the last column.

\renewcommand{\arraystretch}{1}

\begin{table}
\begin{center}
\begin{tabular}{lcccc}
\toprule 
 & {Method} & $ROC_{I}$ & $ROC_{P}$ &  {PRO} \\
\midrule 

\parbox[t]{2mm}{\multirow{2}{*}{\rotatebox[origin=c]{90}{Aug.}}} 
& DRAEM~\cite{zavrtanik2021draem} & 50.9 &  44.0 &  13.6  \\
& SimpleNet~\cite{liu2023simplenet} & 54.9 & 76.1 &  42.4   \\
\midrule 

\parbox[t]{2mm}{\multirow{3}{*}{\rotatebox[origin=c]{90}{Emb.}}} 
& CFA~\cite{lee2022cfa} & 55.7 &  81.3 & 48.8  \\
& CFLOW-AD~\cite{gudovskiy2022cflow} & 77.0 &  94.8 & 80.4 \\
& PyramidalFlow~\cite{lei2023pyramidflow} & 54.4 & 71.1 & 34.9   \\
\midrule 

\parbox[t]{2mm}{\multirow{3}{*}{\rotatebox[origin=c]{90}{Hyb.}}} 
& UniAD~$\dagger$~\cite{uniad} & 83.1  &  97.4 & 87.1  \\
& RD++~\cite{tien2023revisiting} & {83.6} & {97.7} & \underline{90.7}    \\
& DesTSeg~\cite{zhang2023destseg} & 79.3 & 80.3 &  56.1 \\
\midrule 

\parbox[t]{2mm}{\multirow{3}{*}{\rotatebox[origin=c]{90}{Rec.}}} 
& RD~\cite{deng2022anomaly} & 82.7 &  97.2 & 90.0  \\
& ViTAD~$\dagger$~\cite{zhang2023exploring} & 82.7 & 97.2 & 84.8   \\
& MambaAD~\cite{he2024mambaad} & \underline{86.3} & \underline{98.5} & 90.5    \\
\midrule 

\parbox[t]{2mm}{\multirow{3}{*}{\rotatebox[origin=c]{90}{Dif.}}} 
& TransFusion~\cite{Fucka_2024_ECCV} & \textbf{78.6} &  84.2 & 61.6 \\
& DiAD~$\dagger$~\cite{he2024diffusion} & 75.6  &  88.0 & 58.1 \\
& {\ours}~$\dagger$ & 78.0 & \textbf{97.1} & \textbf{87.7} \\

\bottomrule 
\end{tabular}
\end{center}
\vspace{-4mm}
\caption{Results on the large scale industrial AD REAL-IAD~\cite{wang2024real} dataset. $\dagger$: multi-class setting}
\label{tab:real_iad_scores}
\vspace{-5mm}
\end{table}

\subsection{Quantitative Results}
\label{sec:numeric_res}

\textbf{Metrics.} Following the common practice, we used three metrics: Mean Area under the ROC curve $mAU-ROC$~\cite{zavrtanik2021draem} on image and pixel levels, and pixel-level Per-Region-Overlap $PRO$~\cite{Bergmann_Fauser_Sattlegger_Steger_2020} score. Note that for a fair comparison, we use the same evaluation script for all the diffusion models, which consistently gave better results compared to the original ones used in each method.

\noindent\textbf{Real-IAD.} Quantitative results are presented in Table~\ref{tab:real_iad_scores}. Our method significantly outperforms diffusion-based approaches on the pixel-level $ROC_P$ and PRO metrics, and achieves on-par performance compared to state-of-the-art methods. DiAD~\cite{he2024diffusion} suffers from the detection of normal regions as anomalies, and TransFusion~\cite{Fucka_2024_ECCV} misses many anomalous locations. We observe that there are some irregularities in the background of the Real-IAD images, which are sometimes detected as anomalies with DiAD and our method. 
Hence, both of these methods have relatively low image-level scores compared to TransFusion. On the other hand, although TransFusion performs well when the defect is obvious, such as a dent or crack, it fails to detect most of the defect types, such as scratches and discolorations.

\begin{table}
\begin{center}
\begin{tabular}{lcccccccc}
\toprule 
 & {Method} & $ROC_{I}$ & $ROC_{P}$ &  {PRO} \\
\midrule 

\parbox[t]{2mm}{\multirow{2}{*}{\rotatebox[origin=c]{90}{Aug.}}} 
& DRAEM~\cite{zavrtanik2021draem} & 53.5  & 49.9 & 15.3  \\
& SimpleNet~\cite{liu2023simplenet}& 55.6 & 60.2  &26.1   \\
\midrule 

\parbox[t]{2mm}{\multirow{3}{*}{\rotatebox[origin=c]{90}{Emb.}}} 
& CFA~\cite{lee2022cfa} & 56.7  &56.2 &17.9  \\
& CFLOW-AD~\cite{gudovskiy2022cflow} & \underline{67.7} & 76.0  & \underline{47.7} \\
& PyramidalFlow~\cite{lei2023pyramidflow} & 51.6 &50.0  &15.0   \\
\midrule 

\parbox[t]{2mm}{\multirow{3}{*}{\rotatebox[origin=c]{90}{Hyb.}}} 
& UniAD~$\dagger$~\cite{uniad} & 55.2 &   64.6  & 34.3   \\
& RD++~\cite{tien2023revisiting} & 57.5 & 68.2 & 42.2    \\
& DesTSeg~\cite{zhang2023destseg} & 54.4  & 54.5  & 24.4  \\
\midrule 

\parbox[t]{2mm}{\multirow{3}{*}{\rotatebox[origin=c]{90}{Rec.}}} 
& RD~\cite{deng2022anomaly} & 57.6 & 66.5 & 39.8  \\
& ViTAD~$\dagger$~\cite{zhang2023exploring} & 66.9 & \underline{76.2} & 39.1   \\
& MambaAD~\cite{he2024mambaad} & 62.8  & 68.9 & 41.6   \\
\midrule 

\parbox[t]{2mm}{\multirow{3}{*}{\rotatebox[origin=c]{90}{Dif.}}} 
& TransFusion~\cite{Fucka_2024_ECCV} & 58.4 & 57.8 &  6.8 \\
& DiAD~$\dagger$~\cite{he2024diffusion} & 59.0 & 68.1 & 30.8 \\
& {\ours}~$\dagger$ & \textbf{59.1}  & \textbf{69.0}  & \textbf{38.8} \\

\bottomrule 
\end{tabular}
\end{center}
\vspace{-4mm}
\caption{Results on the real world COCO-AD~\cite{coco_ad} dataset.}
\label{tab:coco_ad_scores}
\vspace{-3mm}
\end{table}

\noindent\textbf{COCO-AD.} In order to show the generalization capability of our method to other domains, we experimented with a real-world AD dataset, see Table~\ref{tab:coco_ad_scores}. In comparison to diffusion methods, {\ours} improved the PRO metric by $8$ points, which shows that our method is not bound to any specific domain and could be used in real-world AD as well. {\ours} achieves on par results on all three metrics compared to other state-of-the-art approaches.

\begin{table}[ht]
\begin{center}
\resizebox{1.0\columnwidth}{!}{
\begin{tabular}{ccc|ccc}
 \toprule 
\multirow{3}{*}{Variants} &  \multicolumn{2}{c}{VLM Desc.}   &  \multicolumn{3}{c}{Metrics} \\
 \cmidrule(lr){2-3}  \cmidrule(lr){4-6}
  & Train &  Inference  & $ROC_I$ & $ROC_P$ & $PRO$  \\
\midrule 
DiAD~\cite{he2024diffusion} &  \XSolidBrush & \XSolidBrush  & 75.6      &  88.0 & 58.1      \\
{\ours}  &   \Checkmark & \Checkmark   &  72.6  &  95.9 & 84.0     \\
\midrule
\textbf{{\ours}} &   \Checkmark & \XSolidBrush   & \textbf{78.0} & \textbf{97.1} & \textbf{87.7}    \\
\bottomrule 
\end{tabular}}
\end{center}
\vspace{-4mm}
\caption{Ablation experiments on REAL-IAD~\cite{wang2024real} using VLM descriptions for training, inference or both.}
\label{tab:abl_on_text_prompt}
\vspace{-3mm}
\end{table}

\subsection{Ablation Experiments}
\label{sec:abl_exp}

\begin{table*}
\begin{center}
\resizebox{0.6\textwidth}{!}{
\begin{tabular}{lccc|ccc}
 \toprule 
\multirow{3}{*}{Variants}    &  \multicolumn{3}{c}{Real-IAD} & \multicolumn{3}{c}{COCO-AD} \\
 \cmidrule(lr){2-4}  \cmidrule(lr){5-7} 
 & $ROC_I$ & $ROC_P$  & PRO & $ROC_I$ & $ROC_P$  & PRO  \\
\midrule 
Blip2~\cite{li2023blip}& 76.8 & 96.6 & 86.3 & 59.9 & 68.5 & 38.1  \\
DeepSeekv3-1.3B~\cite{liu2024deepseek} & 77.2  & 97.0 & 87.5  & 58.5 & 68.1 & 38.1  \\
InternVL-2-8B~\cite{captioner} & 78.0 & 97.1 & 87.7 & 59.1 & 69.0 & 38.8   \\
\hline
GT caption  & N/A & N/A   & N/A & 61.1 &  66.5 &   39.9   \\
 \bottomrule 
\end{tabular}}
\end{center}
\vspace{-5mm}
\caption{Comparison of different VLMs for extracting image descriptions. We also use available human captions on COCO-AD. InternVL-2 achieved the best overall performance.}
\label{tab:ablations_vlm}
\vspace{-4mm}
\end{table*}

\textbf{Effect of using text descriptions in inference.} 
Recall that we train our model with only normal images and corresponding textual descriptions extracted using a pretrained VLM. During inference, we experimented with two variants; a) using textual descriptions with the $P_D$ and b) using no text description at all. The results are presented in Table~\ref{tab:abl_on_text_prompt} for Real-IAD dataset, along with the no VLM baseline DiAD~\cite{he2024diffusion}. The test without any textual descriptions achieves the best overall performance. We observed that for some of the test images with a visible defect, the VLM generates descriptions with additional anomaly info such as \textit{horizontal crack present}, \textit{there is a dent on the battery}. These descriptions guide the diffusion model to generate an image with the mentioned defects which harms the performance of the {\ours}. \\
However, on the real image dataset COCO-AD where the anomalies are unseen objects, using textual descriptions boosts the performance by $3$ points on every metric.

\noindent\textbf{Exploring different Vision Language Models.} 
Second, we explore different VLMs to guide the training of the diffusion model, see Table~\ref{tab:ablations_vlm}. 
In addition to InternVL-2, we also extract image descriptions using Blip2~\cite{li2023blip} and the recent DeepSeekVL-v3~\cite{liu2024deepseek} model. We observe that Blip2 generates very short descriptions and fails to describe the objects in detail, especially in the industrial domain datasets. This phenomenon is reflected in the results, and Blip2 achieved the lowest performance. On the other hand, DeepSeekVL and InternVL describe the main object or the image very clearly by focusing on details. On all datasets, these two VLMs achieved similar performance. On the COCO-AD dataset, we also train a model using the available ground truth image captions coming from five different annotators. This model achieved the highest image-level score because the captions provide information about all the normal classes present in the images.

\noindent\textbf{Varying Feature Extractor.} 
We experimented with variants of the DINO~\cite{dino}, DINO-v2~\cite{dino2} and also an ImageNet~\cite{deng2009imagenet} pretrained ResNet~\cite{resnet} model for the segmentation part in Table~\ref{tab:ablations_seg}. 
Using an ImageNet pretrained model as the feature extractor boosts the image-level metric by almost $3$ points. The main reason for this increase stems from the strong confidence scores for the anomaly locations. Using the DINO model with big patch sizes drops the performance on the pixel-level metrics as expected. Since the defects on the industrial datasets are very small, having a single feature for a large image patch suffers dramatically. Note that, for the DINO-v2, the test image is resized to $224\times224$ pixels.

\begin{table}
\begin{center}
\resizebox{0.6\columnwidth}{!}{
\begin{tabular}{c|ccc}
 \toprule 
 {SD AE}   &  \multicolumn{3}{c}{Metrics} \\
  \cmidrule(lr){1-1}  \cmidrule(lr){2-4}
  Train  & $ROC_I$ & $ROC_P$ & $PRO$  \\
\midrule 
 \XSolidBrush  & 77.0     &  95.5 & 83.7      \\
 \Checkmark    & \textbf{78.0} & \textbf{97.1} & \textbf{87.7}    \\
\bottomrule 
\end{tabular}}
\end{center}
\vspace{-3mm}
\caption{The effect of finetuning the image auto-encoder.} 
\label{tab:abl_on_sd_ae}
\vspace{-5mm}
\end{table}

\noindent\textbf{Using pretrained Image Auto-encoder.} 
As an ablation experiment (see Table~\ref{tab:abl_on_sd_ae}), we trained {\ours} on Real-IAD dataset using the baseline Stable Diffusion image auto-encoder weights without any finetuning. Note that in this setting we are training only the denoising UNet. The performance metrics drop significantly, e.g. $ROC_I$ $1.0$, $ROC_P$ $1.6$ and $PRO$ $4.0$ points, which shows the importance of finetuning the image auto-encoder in specialized datasets.

\noindent\textbf{Using only VLM descriptions for AD.} 
As an additional experiment, we use the VLM to determine only the presence or absence of anomalies and evaluate its image-level performance in terms of image-level ROC ($ROC_I$) for each class. On the Real-IAD dataset, the average $ROC_I$ across all classes is $54.5$, which is substantially lower than the performance achieved by our method (cf. Table~\ref{tab:real_iad_scores}). This highlights the limitations of relying solely on VLM responses for anomaly detection and underscores the advantage of our approach in using VLMs for AD. For a more detailed analysis of using VLMs for only anomaly detection, refer to~\cite{xu2025towards}.

\begin{table}
\begin{center}
\resizebox{0.75\columnwidth}{!}{
\begin{tabular}{lccc}
 \toprule 
\multirow{3}{*}{Model}  &  \multicolumn{3}{c}{Metrics}    \\
 \cmidrule(lr){2-4} 
  &  $ROC_I$ & $ROC_P$  & PRO   \\
\midrule 
DINO-v2 ViTS/14 &   61.4    &  58.3   &   16.1   \\
DINO R50        &   65.5    &  88.6   &   56.8    \\
DINO ViTB/16    &   77.2    &  94.6   &   77.5   \\
DINO ViTS/16    &   75.6    &  94.7   &   76.8   \\
DINO ViTB/8     &   76.6    &  96.4   &   85.7  \\
DINO ViTS/8     &   78.0     &  \textbf{97.1}      & \textbf{87.7}   \\
\hline
ImageNet R50    & \textbf{81.1} &  {96.3} & 83.6     \\
 \bottomrule 
\end{tabular}}
\end{center}
\vspace{-3mm}
\caption{Ablation study on using variants of DINO and an ImageNet pretrained ResNet-50 on the Real-IAD.}
\label{tab:ablations_seg}
\vspace{-5mm}
\end{table}

\subsection{Limitations}
\label{sec:limits}

Our current approach utilizes a fixed, simple query to extract image descriptions from VLMs, a design choice made to focus on the core diffusion-based anomaly detection method. While this straightforward strategy already significantly improves upon previous diffusion-based anomaly detection methods, the full potential of VLMs could be further harnessed. Future work could explore optimizing these queries through prompt learning techniques. Additionally, designing queries tailored to the specific characteristics of each dataset could yield richer descriptions and, consequently, lead to enhanced performance.

\section{Conclusion}
\label{sec:conc}

Our proposed framework, {\ours}, effectively bridges the gap between the descriptive power of VLMs and the reconstruction capabilities of diffusion models for multi-class anomaly detection. By leveraging simple, generic prompts and avoiding the limitations of per-class training and synthetic anomaly generation, {\ours} significantly advances the state-of-the-art, achieving substantial performance gains, particularly in pixel-level metrics. Our approach not only enhances generalization, scalability, and efficiency but also opens new avenues for robust and practical anomaly detection in diverse and complex real-world scenarios.

{
    \small
    \bibliographystyle{ieeenat_fullname}
    \bibliography{main}
}

\clearpage
\setcounter{page}{1}
\maketitlesupplementary

\section{Overview}
\label{sec:supp_overview}

In this supplementary material, we first share the quantitative results on MVTec-AD~\cite{bergmann2019mvtec} and VISA~\cite{visa} datasets and comparisons with the same methods presented in the main text. Then, we present per-class results on the Real-IAD~\cite{wang2024real}, and per-split results on the COCO-AD~\cite{coco_ad} dataset, only for the diffusion-based methods to compare. Finally, we show more visual results of our method and diffusion methods on each class of the Real-IAD dataset, and on each split of the COCO-AD dataset.

\begin{table}[h]
    \centering
    \caption{Details of MVTec-AD and VISA datasets.}
    \setlength\tabcolsep{4.0pt}
    \resizebox{1.0\linewidth}{!}{
        \begin{tabular}{cccccccc}
        \toprule
        \multirow{3}{*}{Dataset} & \multicolumn{2}{c}{Categories} & & \multicolumn{4}{c}{Images}  \\
        \cline{2-3} \cline{5-8} 
        & \multirow{2}{*}{Train} & \multirow{2}{*}{Test} & & Train & & \multicolumn{2}{c}{Test} \\
        \cline{5-5} \cline{7-8}
        & & & & Normal & & Anomaly & Normal \\
        \toprule
        MVTec AD~\cite{bergmann2019mvtec}              & 15 &  15 & & 3,629  & & 1,258 & 467   \\ 
        VisA~\cite{visa}                & 12 &	12 & & 8,659  & & 962   & 1,200  \\
        \bottomrule
        \end{tabular}}
    \label{tab:dataset_mvtec_visa}
\end{table}

\begin{table}
\begin{center}
\resizebox{1.0\columnwidth}{!}{
\begin{tabular}{lcccc}
\toprule 
 & {Method} & $ROC_{I}$ & $ROC_{P}$ &  {PRO} \\
\midrule 

\parbox[t]{2mm}{\multirow{3}{*}{\rotatebox[origin=c]{90}{Aug.}}} 
& DRAEM~\cite{zavrtanik2021draem} & 54.5/55.2  & 47.6/48.7 & 14.3/15.8\\
& SimpleNet~\cite{liu2023simplenet} & 95.4/79.2 & 96.8/82.4 & 86.9/62.0  \\
& RealNet~\cite{zhang2024realnet} & 84.8/82.9  & 72.6/69.8  & 56.8/51.2 \\
\midrule 

\parbox[t]{2mm}{\multirow{4}{*}{\rotatebox[origin=c]{90}{Emb.}}} 
& CFA~\cite{lee2022cfa} & 57.6/55.8 & 54.8/43.9 & 25.3/19.3 \\
& PatchCore~\cite{roth2022towards} & \underline{98.8}/ -  & \underline{98.3}/ -  & \underline{94.2} / -  \\
& CFLOW-AD~\cite{gudovskiy2022cflow} & 91.6/92.7  & 95.7/95.8 & 88.3/89.0 \\
& PyramidalFlow~\cite{lei2023pyramidflow} & 70.2/66.2 &   80.0/74.2 & 47.5/40.0   \\
\midrule 

\parbox[t]{2mm}{\multirow{3}{*}{\rotatebox[origin=c]{90}{Hyb.}}} 
& UniAD~$\dagger$~\cite{uniad} & 92.5/96.8 &  95.8/96.8 &  89.3/91.0  \\
& RD++~\cite{tien2023revisiting} & 97.9/95.8 & 97.3/97.3&   93.2/92.9   \\
& DesTSeg~\cite{zhang2023destseg} & 96.4/96.3 & 92.0/92.6 & 83.4/82.6  \\
\midrule 

\parbox[t]{2mm}{\multirow{3}{*}{\rotatebox[origin=c]{90}{Rec.}}} 
& RD~\cite{deng2022anomaly} & 93.6/90.5 & 95.8/95.9 & 91.2/91.2 \\
& ViTAD~$\dagger$~\cite{zhang2023exploring} & 98.3/98.4  & 97.6/97.5 & 92.0/91.7  \\
& MambaAD~\cite{he2024mambaad} & 97.8/98.5 &97.4/97.6 &93.4/93.6  \\
\midrule 

\parbox[t]{2mm}{\multirow{4}{*}{\rotatebox[origin=c]{90}{Dif.}}} 
& DiffAD~\cite{zhang2023unsupervised}  & 80.7/91.8 &  79.7/88.4 & 65.1/78.4 \\
& TransFusion~\cite{Fucka_2024_ECCV}  & 90.4/\textbf{95.3} & 80.9/90.6 & 72.4/83.5 \\
& DiAD~$\dagger$~\cite{he2024diffusion} & 88.9/92.0 & 89.3/89.3 & 63.9/64.4 \\
& \ours~$\dagger$ & 86.9/90.6 & 94.9/\textbf{95.9} & 86.7/\textbf{89.4}\\

\bottomrule 
\end{tabular}}
\end{center}
\caption{Results on the MVTec AD dataset~\cite{bergmann2019mvtec} for 100/300 epochs training. $\dagger$: multi-class setting.}
\label{tab:mvtec_scores_supp}
\end{table}

\begin{table}
\begin{center}
\resizebox{1.0\columnwidth}{!}{
\begin{tabular}{lcccc}
\toprule 
 & {Method} & $ROC_{I}$ & $ROC_{P}$ &  {PRO} \\
\midrule 

\parbox[t]{2mm}{\multirow{3}{*}{\rotatebox[origin=c]{90}{Aug.}}} 
& DRAEM~\cite{zavrtanik2021draem} & 55.1/56.2 & 37.5/45.0 &10.0/16.0\\
& SimpleNet~\cite{liu2023simplenet} & 86.4/80.7&96.6/94.4&79.2/74.2  \\
& RealNet~\cite{zhang2024realnet} & 71.4/79.2&61.0/65.4&27.4/33.9 \\
\midrule 

\parbox[t]{2mm}{\multirow{3}{*}{\rotatebox[origin=c]{90}{Emb.}}} 
& CFA~\cite{lee2022cfa} & 66.3/67.1&81.3/83.0&50.8/48.7 \\
& CFLOW-AD~\cite{gudovskiy2022cflow} & 86.5/87.2&97.7/97.8&86.8/87.3 \\
& PyramidalFlow~\cite{lei2023pyramidflow} & 58.2/69.0&77.0/79.1&42.8/52.6   \\
\midrule 

\parbox[t]{2mm}{\multirow{3}{*}{\rotatebox[origin=c]{90}{Hyb.}}} 
& UniAD~$\dagger$~\cite{uniad} & 89.0/91.4&98.3/\underline{98.5}&86.5/89.0  \\
& RD++~\cite{tien2023revisiting} & \underline{93.9}/93.1&{98.4}/98.4& \underline{91.9}/91.4   \\
& DesTSeg~\cite{zhang2023destseg} & 89.9/89.0&86.7/84.8&61.1/57.5  \\
\midrule 

\parbox[t]{2mm}{\multirow{3}{*}{\rotatebox[origin=c]{90}{Rec.}}} 
& RD~\cite{deng2022anomaly} & 90.6/\underline{93.9} &98.0/98.1& \underline{91.9}/{91.9} \\
& ViTAD~\cite{zhang2023exploring} & 90.4/90.3&98.2/98.2&85.7/85.8  \\
& MambaAD~\cite{he2024mambaad} & 94.5/93.6&98.4/98.2&{92.1}/90.5  \\
\midrule 

\parbox[t]{2mm}{\multirow{4}{*}{\rotatebox[origin=c]{90}{Dif.}}} 
& DiffAD~\cite{zhang2023unsupervised}  & 78.6/89.2 & 82.9/85.5 & 65.7/76.7 \\
& TransFusion~\cite{Fucka_2024_ECCV}  & 87.4/\textbf{92.5}  & 82.1/90.3 & 55.4/64.7 \\
& DiAD~$\dagger$~\cite{he2024diffusion} & 84.8/90.5  & 82.5/83.4 &44.5/44.3 \\
& \ours~$\dagger$ & 79.0/80.9  & 96.0/\textbf{97.0} & 77.0/\textbf{81.0} \\

\bottomrule 
\end{tabular}}
\end{center}
\caption{Results on the VISA AD dataset~\cite{visa} for 100/300 epochs training. $\dagger$: multi-class setting.}
\label{tab:visa_scores_supp}
\end{table}

\subsection{MVTec and VISA results}
\label{sec:supp_mvtec_visa}

Dataset statistics for MVTec-AD~\cite{bergmann2019mvtec} and VISA~\cite{visa} are presented in Table~\ref{tab:dataset_mvtec_visa}.
We trained the best-performing diffusion methods for 100 and 300 epochs on the MVTec-AD~\cite{bergmann2019mvtec} and VISA~\cite{visa} datasets to compare their performance on the same epoch training regime. We present the results in Table~\ref{tab:mvtec_scores_supp} and Table~\ref{tab:visa_scores_supp} for MVTec-AD and VISA, respectively. On MVTec-AD, our method achieved the best $ROC_P$ and PRO scores among the diffusion-based approaches, which show the exceptional localization performance of \ours.
VISA dataset results show similar patterns and our method achieved the best $ROC_P$ score by improving more than $5$ points. 

We conducted extended ablation studies to thoroughly evaluate our method. These experiments focused on three key aspects: 1) the choice of VLMs for extracting image descriptions during training, 2) the impact of including a specific prompt during the inference stage, and 3) the selection of the feature extractor for inference.

Our comparison of VLMs for image description extraction (Table~\ref{tab:ablations_vlm_supp}) revealed that InternVL-2 consistently achieved the best overall performance across both datasets. Further investigation into inference-time prompting (Table~\ref{tab:query_abl_supp}) with InternVL-2 showed that employing prompt $\mathcal{P_D}$ led to a noticeable performance drop on both datasets. Lastly, our analysis of different feature extractors during inference (Table~\ref{tab:ablations_seg_supp}) indicated that DINO ViT-S with a patch size of 8 delivered the strongest overall results.


\subsection{COCO-AD per split results}
\label{sec:supp_coco}
Per-split results on COCO-AD~\cite{coco_ad} are shown in Table~\ref{tab:coco_ad_per_split_scores}. {\ours} shows a noticeable improvement compared to the baselines, especially in the first split where there are significantly fewer normal images.

\subsection{Real-IAD per class results}
\label{sec:supp_real}
Tables~\ref{tab:realiad_class_wise_res1} and~\ref{tab:realiad_class_wise_res2} present per-class results on the Real-IAD~\cite{wang2024real} dataset for diffusion-based methods. Except for a few cases, {\ours} achieves the best $ROC_P$ and $PRO$ on all classes.  A detailed overview of the performance of other methods can be found in \cite{ader}.

\subsection{More visuals for Real-IAD}
\label{sec:supp_vis_real}
\begin{figure*}
\captionsetup[subfigure]{labelformat=empty}
\centering
\resizebox{1.0\textwidth}{!}{
\begin{tabular}{l@{}c@{}c@{}c@{}c@{}c@{}c@{}c@{}c@{}c@{}c}
& \multicolumn{2}{c}{audiojack} & \multicolumn{2}{c}{bottle cap} & \multicolumn{2}{c}{button battery} & \multicolumn{2}{c}{end cap} & \multicolumn{2}{c}{eraser}\\

{\rotatebox[origin=t]{90}{\textit{{I}}}}  & 
\includegraphics[width=\mywidth,  ,valign=m, keepaspectratio,] {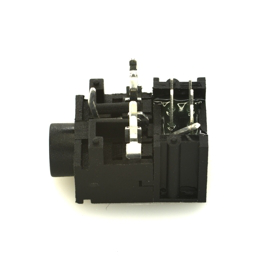}  &
\includegraphics[width=\mywidth,  ,valign=m, keepaspectratio,] {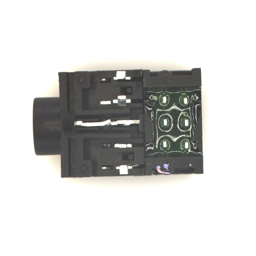} &
\includegraphics[width=\mywidth,  ,valign=m, keepaspectratio,] {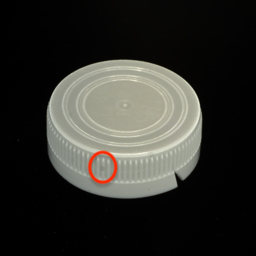} &
\includegraphics[width=\mywidth,  ,valign=m, keepaspectratio,] {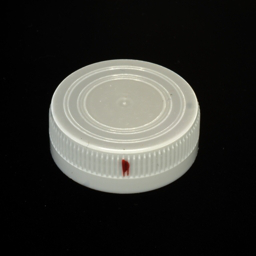} &
\includegraphics[width=\mywidth,  ,valign=m, keepaspectratio,] {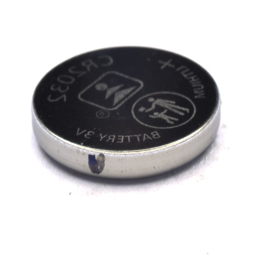} &
\includegraphics[width=\mywidth,  ,valign=m, keepaspectratio,] {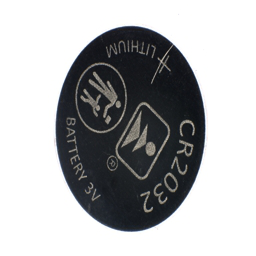} &
\includegraphics[width=\mywidth,  ,valign=m, keepaspectratio,] {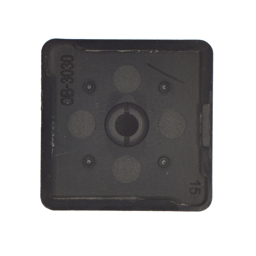} &
\includegraphics[width=\mywidth,  ,valign=m, keepaspectratio,] {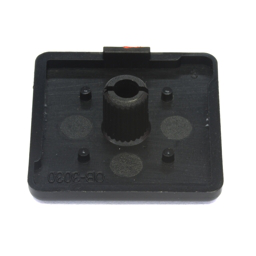} &
\includegraphics[width=\mywidth,  ,valign=m, keepaspectratio,] {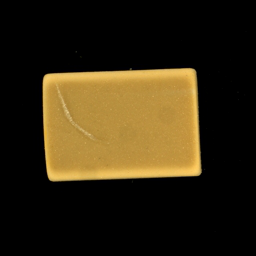} &
\includegraphics[width=\mywidth,  ,valign=m, keepaspectratio,] {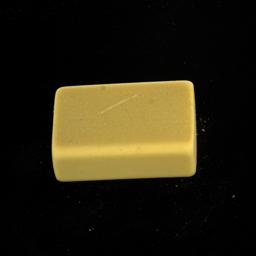}
 \\
\end{tabular}}
\caption{Visual comparison of diffusion-based methods on Real-IAD dataset.}
\label{fig:realiad_comparison1}
\end{figure*}

\begin{figure*}
\captionsetup[subfigure]{labelformat=empty}
\centering
\resizebox{1.0\textwidth}{!}{
\begin{tabular}{l@{}c@{}c@{}c@{}c@{}c@{}c@{}c@{}c@{}c@{}c}
& \multicolumn{2}{c}{fire hood} & \multicolumn{2}{c}{mint} & \multicolumn{2}{c}{mounts} & \multicolumn{2}{c}{pcb} & \multicolumn{2}{c}{phone battery}\\

{\rotatebox[origin=t]{90}{\textit{{I}}}}  & 
\includegraphics[width=\mywidth,  ,valign=m, keepaspectratio,] {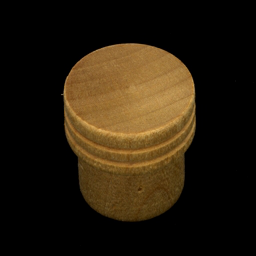}  &
\includegraphics[width=\mywidth,  ,valign=m, keepaspectratio,] {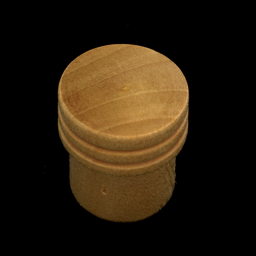} &
\includegraphics[width=\mywidth,  ,valign=m, keepaspectratio,] {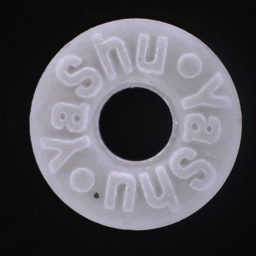} &
\includegraphics[width=\mywidth,  ,valign=m, keepaspectratio,] {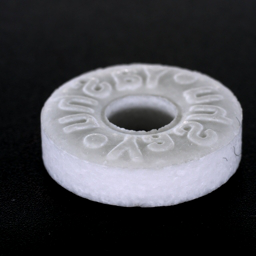} &
\includegraphics[width=\mywidth,  ,valign=m, keepaspectratio,] {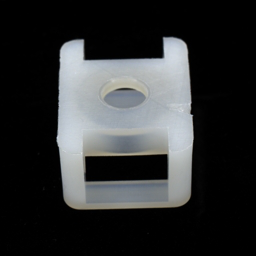} &
\includegraphics[width=\mywidth,  ,valign=m, keepaspectratio,] {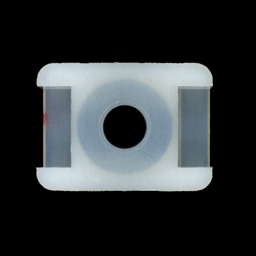} &
\includegraphics[width=\mywidth,  ,valign=m, keepaspectratio,] {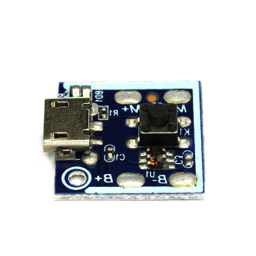} &
\includegraphics[width=\mywidth,  ,valign=m, keepaspectratio,] {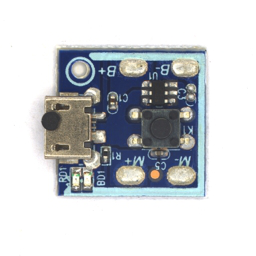} &
\includegraphics[width=\mywidth,  ,valign=m, keepaspectratio,] {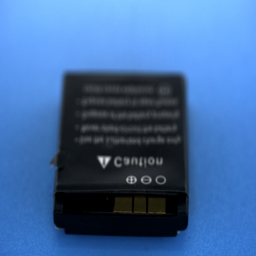} &
\includegraphics[width=\mywidth,  ,valign=m, keepaspectratio,] {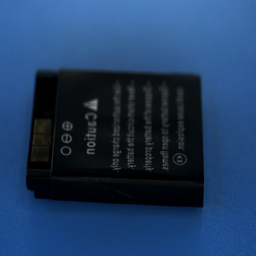} \\
\end{tabular}}
\caption{Visual comparison of diffusion-based methods on Real-IAD dataset.}
\label{fig:realiad_comparison2}
\end{figure*}

\begin{figure*}
\captionsetup[subfigure]{labelformat=empty}
\centering
\resizebox{1.0\textwidth}{!}{
\begin{tabular}{l@{}c@{}c@{}c@{}c@{}c@{}c@{}c@{}c@{}c@{}c}
& \multicolumn{2}{c}{plastic nut} & \multicolumn{2}{c}{plastic plug} & \multicolumn{2}{c}{porcelain doll} & \multicolumn{2}{c}{regulator} & \multicolumn{2}{c}{rolled strip} \\

{\rotatebox[origin=t]{90}{\textit{{I}}}}  & 
\includegraphics[width=\mywidth,  ,valign=m, keepaspectratio,] {./figures/vis_comp/realiad/plastic_nut_0028_NG_ZW_C4_20231004094025.jpg_true.png}  &
\includegraphics[width=\mywidth,  ,valign=m, keepaspectratio,] {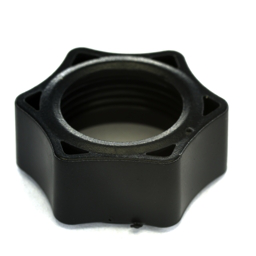} &
\includegraphics[width=\mywidth,  ,valign=m, keepaspectratio,] {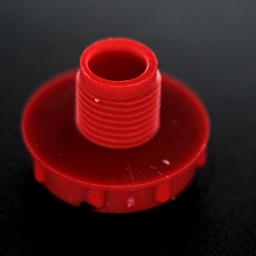}  &
\includegraphics[width=\mywidth,  ,valign=m, keepaspectratio,] {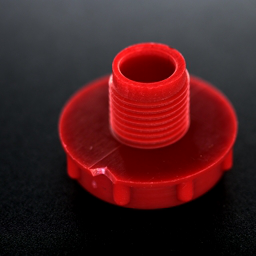} &
\includegraphics[width=\mywidth,  ,valign=m, keepaspectratio,] {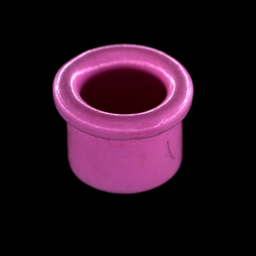} &
\includegraphics[width=\mywidth,  ,valign=m, keepaspectratio,] {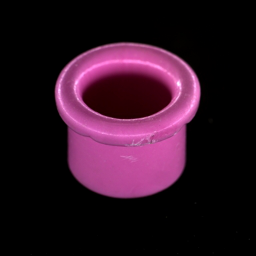} &
\includegraphics[width=\mywidth,  ,valign=m, keepaspectratio,] {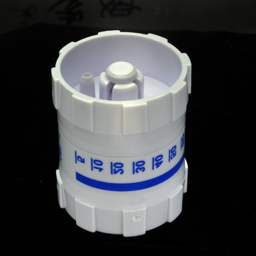} &
\includegraphics[width=\mywidth,  ,valign=m, keepaspectratio,] {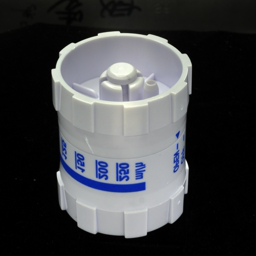} &
\includegraphics[width=\mywidth,  ,valign=m, keepaspectratio,] {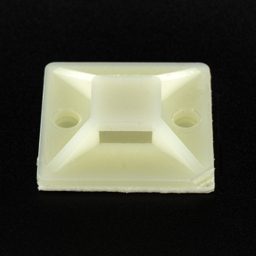} &
\includegraphics[width=\mywidth,  ,valign=m, keepaspectratio,] {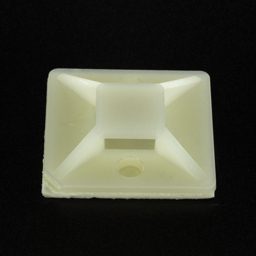} \\
\end{tabular}}
\caption{Visual comparison of diffusion-based methods on Real-IAD dataset.}
\label{fig:realiad_comparison3}
\end{figure*}

\begin{figure*}
\captionsetup[subfigure]{labelformat=empty}
\centering
\resizebox{1.0\textwidth}{!}{
\begin{tabular}{l@{}c@{}c@{}c@{}c@{}c@{}c@{}c@{}c@{}c@{}c}
& \multicolumn{2}{c}{sim card} & \multicolumn{2}{c}{switch} & \multicolumn{2}{c}{tape} & \multicolumn{2}{c}{terminal block} & \multicolumn{2}{c}{toothbrush} \\

{\rotatebox[origin=t]{90}{\textit{{I}}}}  & 
\includegraphics[width=\mywidth,  ,valign=m, keepaspectratio,] {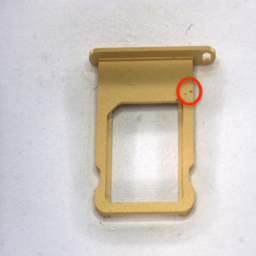} &
\includegraphics[width=\mywidth,  ,valign=m, keepaspectratio,] {./figures/vis_comp/realiad/sim_card_set_0006_NG_ZW_C3_20230923091153.jpg_true.png} &
\includegraphics[width=\mywidth,  ,valign=m, keepaspectratio,] {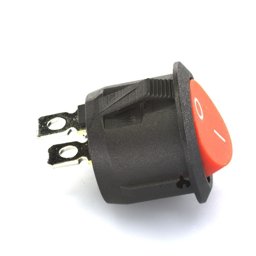}  &
\includegraphics[width=\mywidth,  ,valign=m, keepaspectratio,] {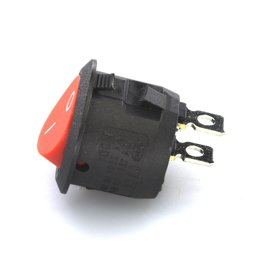} &
\includegraphics[width=\mywidth,  ,valign=m, keepaspectratio,] {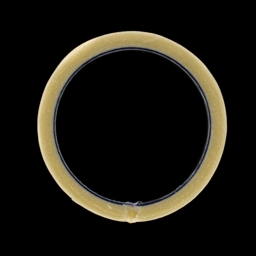} &
\includegraphics[width=\mywidth,  ,valign=m, keepaspectratio,] {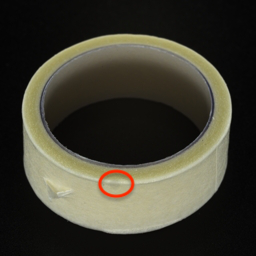} &
\includegraphics[width=\mywidth,  ,valign=m, keepaspectratio,] {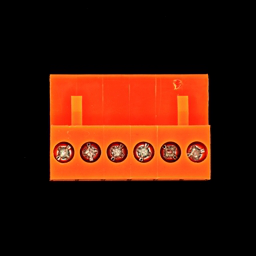} &
\includegraphics[width=\mywidth,  ,valign=m, keepaspectratio,] {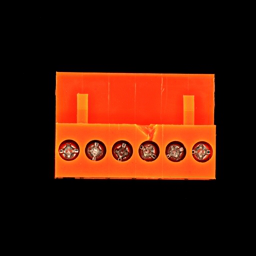} &
\includegraphics[width=\mywidth,  ,valign=m, keepaspectratio,] {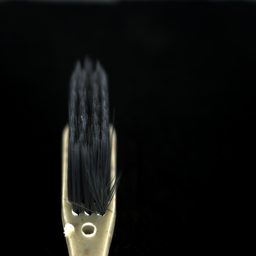} &
\includegraphics[width=\mywidth,  ,valign=m, keepaspectratio,] {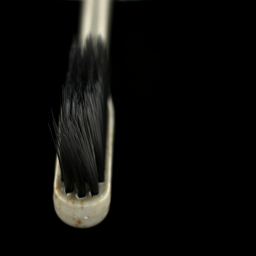} \\
\end{tabular}}
\caption{Visual comparison of diffusion-based methods on Real-IAD dataset.}
\label{fig:realiad_comparison4}
\end{figure*}

\begin{figure*}
\captionsetup[subfigure]{labelformat=empty}
\centering
\resizebox{1.0\textwidth}{!}{
\begin{tabular}{l@{}c@{}c@{}c@{}c@{}c@{}c@{}c@{}c@{}c@{}c}
& \multicolumn{2}{c}{toy} & \multicolumn{2}{c}{toy brick} & \multicolumn{2}{c}{transistor} & \multicolumn{2}{c}{u block} & \multicolumn{2}{c}{usb} \\

{\rotatebox[origin=t]{90}{\textit{{I}}}}  & 
\includegraphics[width=\mywidth,  ,valign=m, keepaspectratio,] {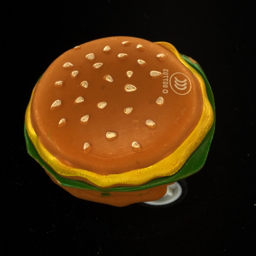} &
\includegraphics[width=\mywidth,  ,valign=m, keepaspectratio,] {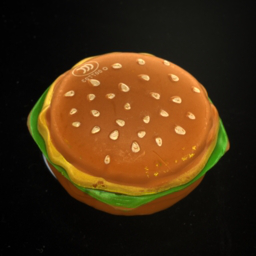} &
\includegraphics[width=\mywidth,  ,valign=m, keepaspectratio,] {./figures/vis_comp/realiad/toy_brick_0001_NG_AK_C1_20230917114449.jpg_true.png}  &
\includegraphics[width=\mywidth,  ,valign=m, keepaspectratio,] {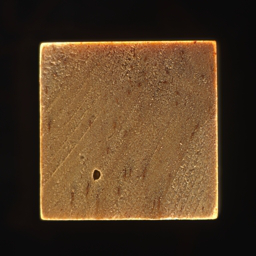} &
\includegraphics[width=\mywidth,  ,valign=m, keepaspectratio,] {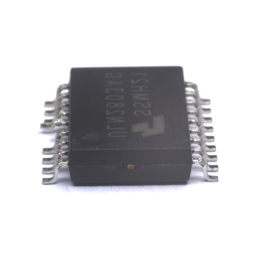} &
\includegraphics[width=\mywidth,  ,valign=m, keepaspectratio,] {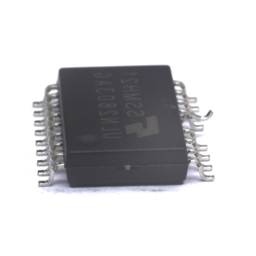} &
\includegraphics[width=\mywidth,  ,valign=m, keepaspectratio,] {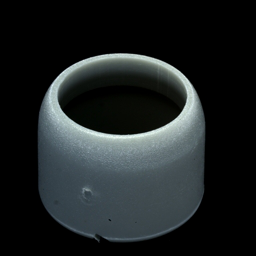} &
\includegraphics[width=\mywidth,  ,valign=m, keepaspectratio,] {./figures/vis_comp/realiad/u_block_0001_NG_ZW_C2_20231005145842.jpg_true.png} &
\includegraphics[width=\mywidth,  ,valign=m, keepaspectratio,] {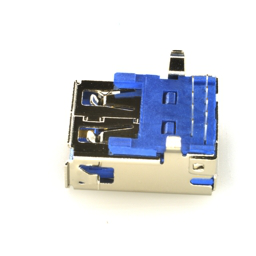} &
\includegraphics[width=\mywidth,  ,valign=m, keepaspectratio,] {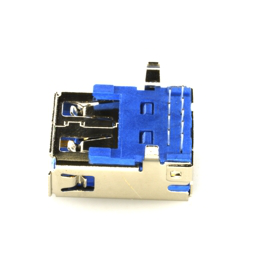} \\
\end{tabular}}
\caption{Visual comparison of diffusion-based methods on Real-IAD dataset.}
\label{fig:realiad_comparison5}
\end{figure*}

\begin{figure*}
\captionsetup[subfigure]{labelformat=empty}
\centering
\resizebox{1.0\textwidth}{!}{
\begin{tabular}{l@{}c@{}c@{}c@{}c@{}c@{}c@{}c@{}c@{}c@{}c}
& \multicolumn{2}{c}{usb adaptor} & \multicolumn{2}{c}{vcpill} & \multicolumn{2}{c}{wooden beads} & \multicolumn{2}{c}{woodstick} & \multicolumn{2}{c}{zipper}  \\

{\rotatebox[origin=t]{90}{\textit{{I}}}}  & 
\includegraphics[width=\mywidth,  ,valign=m, keepaspectratio,] {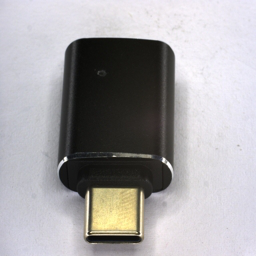} &
\includegraphics[width=\mywidth,  ,valign=m, keepaspectratio,] {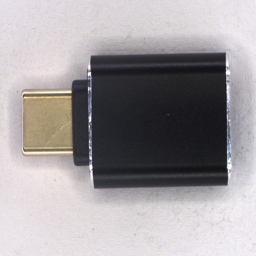} &
\includegraphics[width=\mywidth,  ,valign=m, keepaspectratio,] {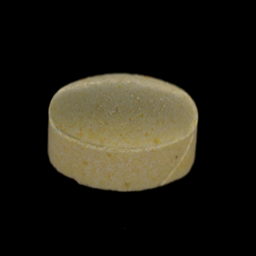}  &
\includegraphics[width=\mywidth,  ,valign=m, keepaspectratio,] {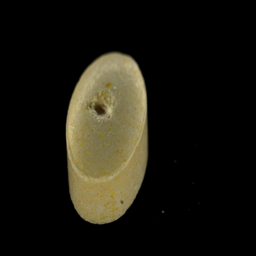} &
\includegraphics[width=\mywidth,  ,valign=m, keepaspectratio,] {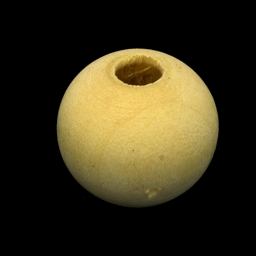} &
\includegraphics[width=\mywidth,  ,valign=m, keepaspectratio,] {./figures/vis_comp/realiad/wooden_beads_0002_NG_HS_C3_20231014093540.jpg_true.png} &
\includegraphics[width=\mywidth,  ,valign=m, keepaspectratio,] {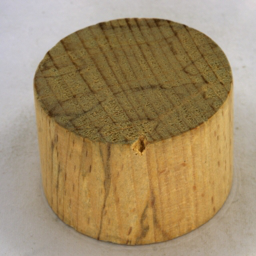} &
\includegraphics[width=\mywidth,  ,valign=m, keepaspectratio,] {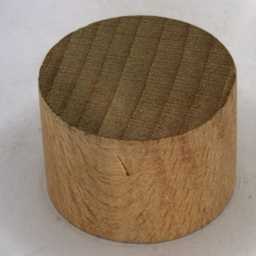} &
\includegraphics[width=\mywidth,  ,valign=m, keepaspectratio,] {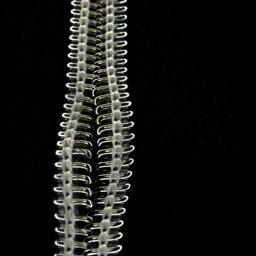} &
\includegraphics[width=\mywidth,  ,valign=m, keepaspectratio,] {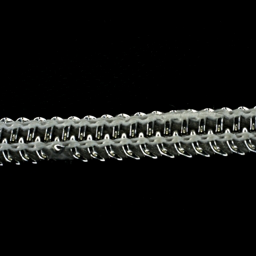} \\
\end{tabular}}
\caption{Visual comparison of diffusion-based methods on Real-IAD dataset.}
\label{fig:realiad_comparison6}
\end{figure*}

We present more visual comparisons in Figures~\ref{fig:realiad_comparison1}-\ref{fig:realiad_comparison6} on Real-IAD. Specifically, we show two results per object category in the dataset. {\ours} shows superior localization capability compared to strong baselines. Moreover, in some cases, we observe that it even finds unmarked potential defective pixels. For instance, in Figure~\ref{fig:realiad_comparison1} first \textit{bottle cap} image has a small blue dot which is marked as an anomaly by our method. Similarly, both \textit{sim card} objects have small defective pixels which are again detected by {\ours}.

\begin{table}
\begin{center}
\resizebox{1.0\columnwidth}{!}{
\begin{tabular}{lccc|ccc}
 \toprule 
\multirow{3}{*}{Variants}  &  \multicolumn{3}{c}{MVTec-AD} &  \multicolumn{3}{c}{VISA}    \\
 \cmidrule(lr){2-4}  \cmidrule(lr){5-7} 
&  $ROC_I$ & $ROC_P$  & PRO &  $ROC_I$ & $ROC_P$  & PRO  \\
\midrule 
Blip2 & 89.8 & 95.7 & 88.6 & 82.3 & 97.0 & 80.7   \\
DeepSeekv3-1.3B & 90.7  & 95.5  & 89.0 & 81.8 & 96.8 & 81.0  \\
InternVL-2-8B & 90.6 & 95.9  & 89.4 & 80.9 & 97.0 & 81.0   \\
 \bottomrule 
\end{tabular}}
\end{center}
\caption{Ablation experiments using different VLMs to extract anomaly descriptions on MVTec-AD and VISA datasets.}
\label{tab:ablations_vlm_supp}
\end{table}

\begin{table}
\begin{center}
\begin{tabular}{lccc}
\toprule 
Dataset & $ROC_{I}$ & $ROC_{P}$ & PRO \\
\midrule 
MVTec-AD & -2.2  & -2.4 & -6.5 \\
VISA  & -8.1 & -2.9 & -9.9 \\
\bottomrule 
\end{tabular}
\end{center}
\caption{Relative change when we use $\mathcal{P_D}$ query during inference. Using text description from InternVL-2 for inference has a negative impact on all metrics.}
\label{tab:query_abl_supp}
\end{table}

\begin{table}
\begin{center}
\resizebox{0.8\columnwidth}{!}{
\begin{tabular}{lccc}
 \toprule 
\multirow{3}{*}{Model}  &  \multicolumn{3}{c}{Metrics}    \\
 \cmidrule(lr){2-4} 
  &  $ROC_I$ & $ROC_P$  & PRO   \\
\midrule 
ImageNet R50    & 88.8 &  92.6 & 81.0    \\
\hline
DINO-v2 ViTS/14 &   64.6   &   61.3   &  16.8    \\
DINO R50        &   86.8    &   90.3   &  71.0    \\
DINO ViTB/16    &   89.8    &  92.8    &  79.3    \\
DINO ViTS/16    &   85.1   &  90.2    & 74.3     \\
DINO ViTB/8     &  87.5 &  93.6    &  82.6   \\
DINO ViTS/8     &   \textbf{90.6}  &  \textbf{95.9}    & \textbf{89.4}   \\
 \bottomrule 
\end{tabular}}
\end{center}
\caption{Ablation experiments using variants of DINO and an ImageNet pretrained Resnet-50 for anomaly segmentation on MVTec-AD dataset with InternVL-2 descriptions.}
\label{tab:ablations_seg_supp}
\end{table}

\subsection{More visuals for COCO-AD}
\label{sec:supp_vis_coco}

\begin{figure*}
\captionsetup[subfigure]{labelformat=empty}
\centering
\resizebox{1.0\textwidth}{!}{
\begin{tabular}{l@{~}c@{~}c@{~}c|c@{~}c@{~}c}
& \multicolumn{3}{c}{{Split-0}} & \multicolumn{3}{c}{{Split-1}}  \\

{\rotatebox[origin=t]{90}{\textit{{\tiny I}}}}  & 
\includegraphics[width=\mywidth,  ,valign=m, keepaspectratio,] {./figures/vis_comp/coco/part0/000000007281.jpg_true.png}  &
\includegraphics[width=\mywidth,  ,valign=m, keepaspectratio,] {./figures/vis_comp/coco/part0/000000008844.jpg_true.png} &
\includegraphics[width=\mywidth,  ,valign=m, keepaspectratio,] {./figures/vis_comp/coco/part0/000000018737.jpg_true.png} &
\includegraphics[width=\mywidth,  ,valign=m, keepaspectratio,] {./figures/vis_comp/coco/part1/000000581100.jpg_true.png}&
\includegraphics[width=\mywidth,  ,valign=m, keepaspectratio,] {./figures/vis_comp/coco/part1/000000010764.jpg_true.png}&
\includegraphics[width=\mywidth,  ,valign=m, keepaspectratio,] {./figures/vis_comp/coco/part1/000000006763.jpg_true.png}\\

{\rotatebox[origin=t]{90}{\textit{\textbf{ \tiny DiAD}}}} &  
\includegraphics[width=\mywidth,  ,valign=m, keepaspectratio,] {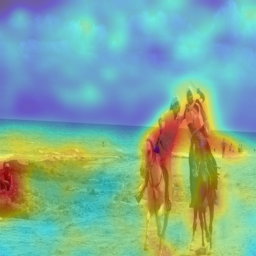}  &
\includegraphics[width=\mywidth,  ,valign=m, keepaspectratio,] {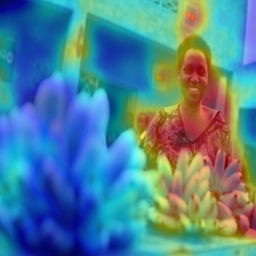} &
\includegraphics[width=\mywidth,  ,valign=m, keepaspectratio,] {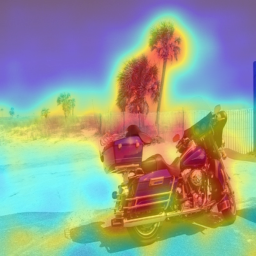} &
\includegraphics[width=\mywidth,  ,valign=m, keepaspectratio,] {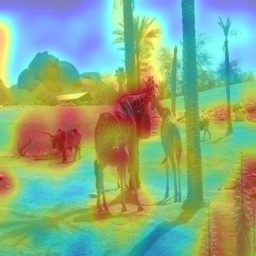} &
\includegraphics[width=\mywidth,  ,valign=m, keepaspectratio,] {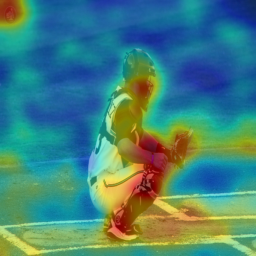} &
\includegraphics[width=\mywidth,  ,valign=m, keepaspectratio,] {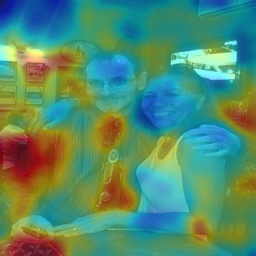} \\
\end{tabular}}
\caption{Visual comparison of diffusion-based methods on COCO-AD dataset on Split-0 and Split-1. }
\label{fig:cocoad_comparison_supp}
\end{figure*}

\begin{figure*}
\captionsetup[subfigure]{labelformat=empty}
\centering
\resizebox{1.0\textwidth}{!}{
\begin{tabular}{l@{~}c@{~}c@{~}c|c@{~}c@{~}c}
& \multicolumn{3}{c}{{Split-2}}  & \multicolumn{3}{c}{{Split-3}}   \\

{\rotatebox[origin=t]{90}{\textit{{\tiny I}}}}  &
\includegraphics[width=\mywidth,  ,valign=m, keepaspectratio,] {./figures/vis_comp/coco/part2/000000473869.jpg_true.png}&
\includegraphics[width=\mywidth,  ,valign=m, keepaspectratio,] {./figures/vis_comp/coco/part2/000000000632.jpg_true.png}&
\includegraphics[width=\mywidth,  ,valign=m, keepaspectratio,] {./figures/vis_comp/coco/part2/000000125778.jpg_true.png}&
\includegraphics[width=\mywidth,  ,valign=m, keepaspectratio,] {./figures/vis_comp/coco/part3/000000018380.jpg_true.png}&
\includegraphics[width=\mywidth,  ,valign=m, keepaspectratio,] {./figures/vis_comp/coco/part3/000000051610.jpg_true.png}&
\includegraphics[width=\mywidth,  ,valign=m, keepaspectratio,] {./figures/vis_comp/coco/part3/000000009400.jpg_true.png} \\

{\rotatebox[origin=t]{90}{\textit{\textbf{ \tiny DiAD}}}} &
\includegraphics[width=\mywidth,  ,valign=m, keepaspectratio,] {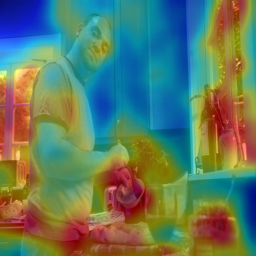} &
\includegraphics[width=\mywidth,  ,valign=m, keepaspectratio,] {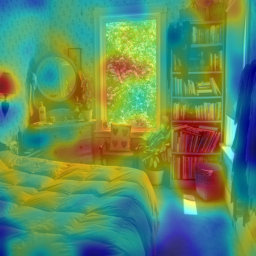} &
\includegraphics[width=\mywidth,  ,valign=m, keepaspectratio,] {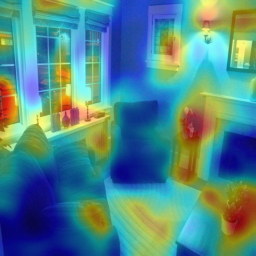} &
\includegraphics[width=\mywidth,  ,valign=m, keepaspectratio,] {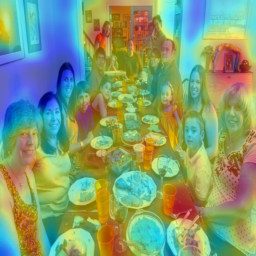} &
\includegraphics[width=\mywidth,  ,valign=m, keepaspectratio,] {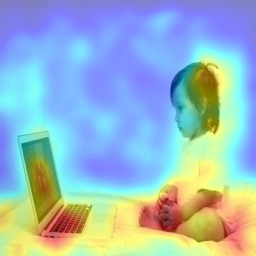} &
\includegraphics[width=\mywidth,  ,valign=m, keepaspectratio,] {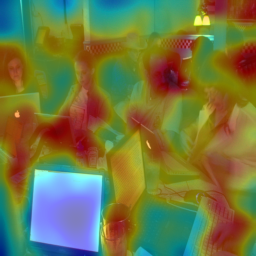} \\
\end{tabular}}
\caption{Visual comparison of diffusion-based methods on COCO-AD dataset on Split-2 and Split-3. }
\label{fig:cocoad_comparison_supp2}
\end{figure*}

Figures~\ref{fig:cocoad_comparison_supp} and \ref{fig:cocoad_comparison_supp2} show three example results per split, and in each split, we pick different anomaly classes to show the performance across various objects. As a real-world dataset, COCO-AD is more complex and challenging compared to previous industrial domain datasets. Nevertheless, \ours achieves significantly better anomaly localization across multiple classes.

\begin{table}
\begin{center}
\begin{tabular}{lcccc}
\toprule 
 & {Method} & $ROC_{I}$ & $ROC_{P}$ &  {PRO} \\
\midrule 
 \multirow{3}{*}{\rotatebox{90}{{Split-0}}} 
& DiAD~$\dagger$ & 57.5 & 67.0 & 28.8 \\
& TransFusion & 56.1 &  54.8 &  12.8 \\
& \ours~$\dagger$ & \textbf{62.6} & \textbf{74.3} & \textbf{43.8}  \\

\hline \multirow{3}{*}{\rotatebox{90}{{Split-1}}} 
& DiAD~$\dagger$ & 54.4 &  71.3 & 28.8 \\
& TransFusion  & \textbf{57.1} & 62.2 & 6.6 \\
& \ours~$\dagger$ & 52.7 & \textbf{71.5} & \textbf{37.5}  \\

\hline \multirow{3}{*}{\rotatebox{90}{{Split-2}}} 
& DiAD~$\dagger$ & \textbf{63.8} & 68.0 & 33.2 \\
& TransFusion  & 61.4 &  58.4 & 2.7 \\
& \ours~$\dagger$ & 62.9 & \textbf{69.3} & \textbf{40.7}  \\

\hline \multirow{3}{*}{\rotatebox{90}{{Split-3}}} 
& DiAD~$\dagger$ & \textbf{60.1} &  \textbf{65.9} &  32.3 \\
& TransFusion  & 59.0 & 56.1 & 5.0 \\
& \ours~$\dagger$ & 58.2 & 60.8 & \textbf{33.2}  \\

\hline \multirow{3}{*}{\rotatebox{90}{{Avg}}} 
& DiAD~$\dagger$ & 59.0 & 68.1 & 30.8 \\
& TransFusion & 58.4 &  57.8 &  6.8 \\
& \ours~$\dagger$ & \textbf{59.1} & \textbf{69.0}  & \textbf{38.8} \\

\bottomrule 
\end{tabular}
\end{center}
\caption{Per split results on the COCO-AD dataset~\cite{coco_ad} for 100 epochs training. $\dagger$: multi-class setting.}
\label{tab:coco_ad_per_split_scores}
\end{table}


\begin{table}[hp]
\centering
\renewcommand{\arraystretch}{1.0}
\setlength\tabcolsep{5.0pt}
\begin{tabular}{lcccc}
\toprule 
 & {Method} & $ROC_{I}$ & $ROC_{P}$ &  {PRO} \\
\midrule 

\multirow{3}{*}{\rotatebox{90}{\tiny{audiojack}}}   
& DiAD & 76.5  &  91.6  & 63.3  \\
& Transfusion & \textbf{80.3}  & 85.9  & 51.2  \\
& \ours & 77.5 &  \textbf{97.8} & \textbf{87.1}  \\

\hline \multirow{3}{*}{\rotatebox{90}{\tiny{{bottle cap}}}} 
& DiAD & \textbf{91.6} &   94.6 & 73.0 \\
& Transfusion & 65.4  & 70.9 & 43.3  \\
& \ours & 77.3 & \textbf{98.4}  & \textbf{92.3}\\

\hline \multirow{3}{*}{\rotatebox{90}{\tiny{button battery}}} 
& DiAD & 80.5 &  84.1 &  66.9  \\
\bottomrule
\end{tabular}
    \caption{Per class results on Real-IAD dataset for diffusion models, part 1.}
    \label{tab:realiad_class_wise_res1}
\end{table}


\begin{table}[hp]
\centering
\renewcommand{\arraystretch}{1.0}
\setlength\tabcolsep{5.0pt}
\begin{tabular}{lcccc}
\toprule 
 & {Method} & $ROC_{I}$ & $ROC_{P}$ &  {PRO} \\
\midrule 

\multirow{3}{*}{\rotatebox{90}{\tiny{sim card}}} 
& DiAD & 72.6 &  89.9 &  60.4  \\
& Transfusion & 91.8  & 96.6 & 82.5  \\
& \ours & \textbf{92.9} & \textbf{98.3}  & \textbf{90.0}  \\

\hline \multirow{3}{*}{\rotatebox{90}{\tiny{switch}}} 
& DiAD & 73.4 &   90.5 &  64.2  \\
& Transfusion & 82.0 & 86.6 &  59.4  \\
& \ours & \textbf{83.9} & \textbf{96.8} & \textbf{90.7}  \\

\hline \multirow{3}{*}{\rotatebox{90}{\tiny{tape}}} 
& DiAD & 73.9 &  81.7 &  47.3 \\
\bottomrule
\end{tabular}
    \caption{Per class results on Real-IAD dataset for diffusion models, part 2.}
    \label{tab:realiad_class_wise_res2}
\end{table}

\clearpage

\renewcommand{\arraystretch}{1}

\renewcommand{\arraystretch}{1}

\renewcommand{\arraystretch}{1}

\renewcommand{\arraystretch}{1}

\renewcommand{\arraystretch}{1}

\renewcommand{\arraystretch}{1}

\renewcommand{\arraystretch}{1}

\renewcommand{\arraystretch}{1}



\end{document}